\documentclass[10pt,twocolumn,letterpaper]{article}

 \usepackage[pagenumbers]{cvpr} 

\usepackage{stfloats}
\usepackage{graphicx}
\usepackage{amsmath}
\usepackage{amssymb}
\usepackage{bbding}
\usepackage{booktabs}
\usepackage{makecell}
\usepackage[export]{adjustbox}
\usepackage[normalem]{ulem}
\usepackage{textpos}  %
\usepackage[table]{xcolor}

\usepackage{tabulary,multirow,overpic,xcolor,subfloat}

\definecolor{cvprblue}{rgb}{0.21,0.49,0.74}
\definecolor{linkcolor}{HTML}{ED1C24}
\usepackage[pagebackref,breaklinks,colorlinks,citecolor=cvprblue, linkcolor=linkcolor]{hyperref}

\newcommand{\app}{\raise.17ex\hbox{$\scriptstyle\sim$}}

\newcolumntype{x}[1]{>{\centering\arraybackslash}p{#1pt}}
\newcolumntype{y}[1]{>{\raggedright\arraybackslash}p{#1pt}}

\newlength\savewidth\newcommand\shline{\noalign{\global\savewidth\arrayrulewidth
		\global\arrayrulewidth 1pt}\hline\noalign{\global\arrayrulewidth\savewidth}}
\newcommand{\tablestyle}[2]{\setlength{\tabcolsep}{#1}\renewcommand{\arraystretch}{#2}\centering\footnotesize}

\newcommand{\figref}[1]{Fig.~\ref{#1}}
\newcommand{\secref}[1]{Sec.~\ref{#1}}
\newcommand{\tabref}[1]{Table~\ref{#1}}
\newcommand{\appref}[1]{Appendix~\ref{#1}}
\newcommand{\equref}[1]{Eq.~\ref{#1}}

\newcommand{\modelname}{UniPortrait\xspace}

\definecolor{Gray}{gray}{0.5}
\newcommand{\demph}[1]{\textcolor{Gray}{#1}}

\newcommand{\gr}{\rowcolor[gray]{.95}}

\def\paperID{*****} 
\def\confName{CVPR}
\def\confYear{2024}

\title{\vskip -0.5em UniPortrait: A Unified Framework for Identity-Preserving Single- and Multi-Human Image Personalization\vspace{-0.5em}}

\vspace{-1em}
\author{Junjie He, Yifeng Geng, and Liefeng Bo\\
	Institute for Intelligent Computing, Alibaba Group\\
{\tt\small \{hejunjie.hjj, cangyu.gyf, liefeng.bo\}@alibaba-inc.com}
}

\begin{document}
	
\makeatletter
\let\@oldmaketitle\@maketitle%
\renewcommand{\@maketitle}{\@oldmaketitle%
	\vskip -2.0em
	\centering
	\includegraphics[width=1\textwidth]{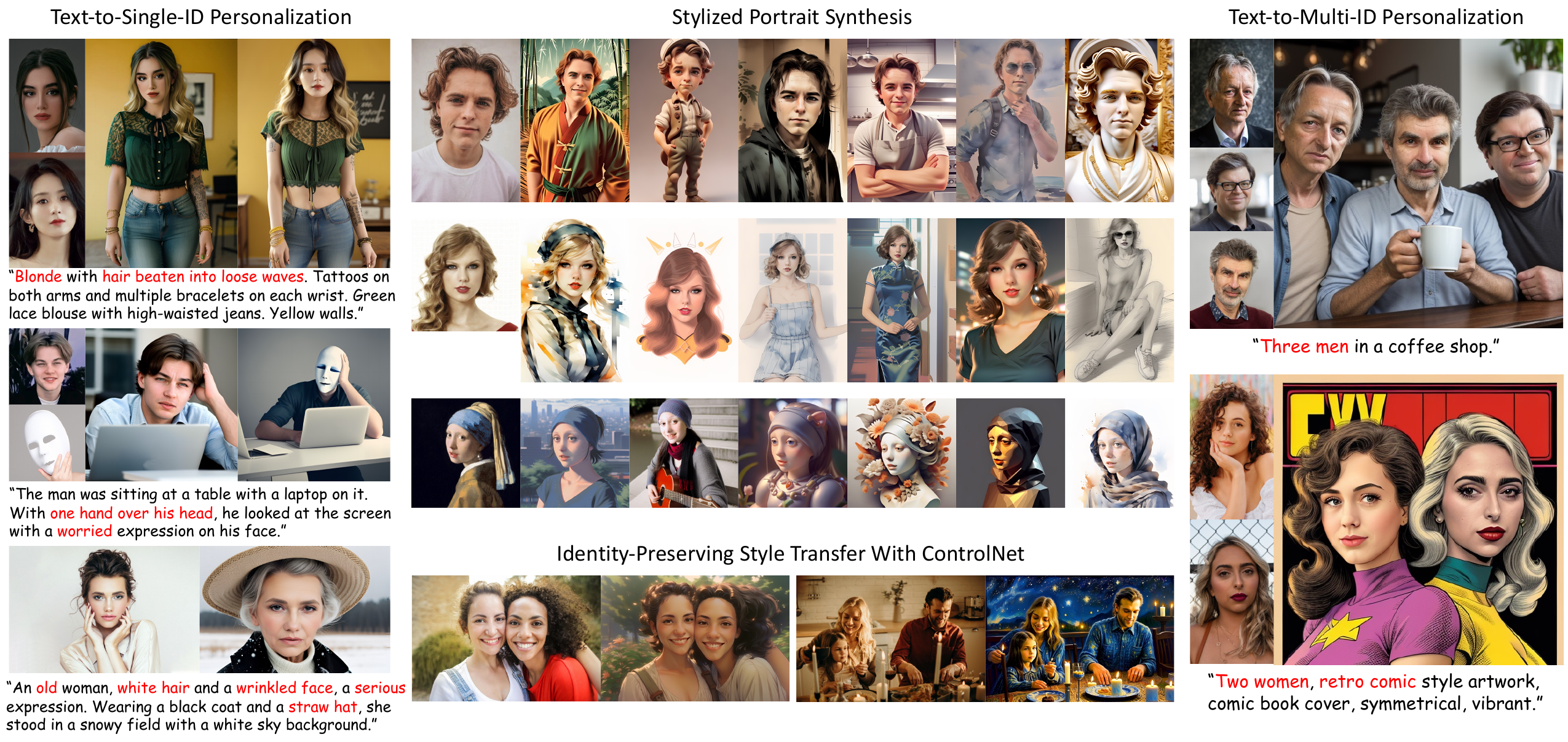}
	\vskip -0.5em
	\captionof{figure}{ 
		\textbf{Example generations from UniPortrait.}  Our method customizes single- and multi-ID images in a unified manner, providing high-fidelity identity preservation, extensive facial editability, free-form text description, and no requirement for a predetermined layout.
	}
	\label{fig:highlight}
	\bigskip}
\makeatother

\maketitle

\begin{abstract}
	
This paper presents UniPortrait, an innovative human image personalization framework that unifies single- and multi-ID customization with high face fidelity, extensive facial editability, free-form input description, and diverse layout generation. UniPortrait consists of only two plug-and-play modules: an ID embedding module and an ID routing module. The ID embedding module extracts versatile editable facial features with a decoupling strategy for each ID and embeds them into the context space of diffusion models. The ID routing module then combines and distributes these embeddings adaptively to their respective regions within the synthesized image, achieving the customization of single and multiple IDs.
With a carefully designed two-stage training scheme, UniPortrait achieves superior performance in both single- and multi-ID customization. Quantitative and qualitative experiments demonstrate the advantages of our method over existing approaches as well as its good scalability, \eg, the universal compatibility with existing generative control tools. The project page is at \url{https://aigcdesigngroup.github.io/UniPortrait-Page/}.
\end{abstract}

\section{Introduction}
\label{sec:intro}

Recent advances in diffusion models~\cite{dhariwal2021diffusion, ho2020denoising} have revolutionized the field of text-to-image synthesis, paving the way for a plethora of image customization tasks~\cite{kumari2022customdiffusion,wei2023elite,li2024blipdiffusion,wang2023styleadapter,subject-diffusion}. Among these, human image personalization has emerged as a key area of focus for its enormous potential in applications such as AI portrait photos, image animation, and virtual try-ons. The objective of this task is to generate images that maintain a consistent face identity (ID) with reference face images while adhering to additional prompts.

Early human image customization methods~\cite{gal2022image,hu2022lora,ruiz2023hyperdreambooth} rely on test-time fine-tuning, with each new identity roughly taking dozens of minutes or even hours to achieve satisfactory results.
Recent works~\cite{li2024photomaker,xiao2023fastcomposer,ye2023ip-adapter,wang2024instantid} resort to tuning-free setups. They employ a separate trained encoder to encode face ID information into one or several tokens and inject them into the generation process, resembling text conditioning. 
Thanks to the elimination of training requirements, these tuning-free methods have achieved impressive efficiency when facing new identities.
However, they still suffer from two limitations. First, they struggle to preserve facial shape and texture details. This is because the features they use are either losing the spatial information (\eg, the global embedding of the encoder~\cite{yan2023facestudio,wang2024instantid}) or not expert at the face domain (\eg the features from CLIP~\cite{radford2021learning} image encoder~\cite{li2024photomaker, fastcomposer}). Second, these encoder-based methods often tend towards mere replication of the reference face image, presenting difficulties in facial control and editing. 
We attribute this problem to the current methods' insufficient disentanglement between intrinsic face identity and face-relevant yet identity-irrelevant representations. As a result, the model overfits the spurious facial details.

Besides, previous human customization methods primarily focus on single-ID personalization and struggle when personalizing multi-ID images. The main challenge they encountered is identity blending~\cite{fastcomposer}, in which a same generated face attends to multiple ID features. Some multi-concept customization methods~\cite{fastcomposer,kumari2023multi,avrahami2023break,jang2024identity} integrate ID information into text embeddings and use the structure of text to distinguish different subjects. 
Nonetheless, these techniques necessitate a one-to-one mapping between identities and corresponding text tokens, preventing the use of free-form text prompts for subjects like plural phrases ``two women''. 
Moreover,  such direct fusion may compromise the integrity of both identity representation and text semantics because of the distinct nature of visual and textual signals.
Another line of research~\cite{liu2023cones,kim2024instantfamily,kwon2024concept,dahary2024yourself,kong2024omg,gu2024mix} introduces manually designed masks to separate the different IDs' information. Despite their effectiveness in avoiding ID blending, they require specifying the face location before the generation, which limits the diversity of generated images.

To address the aforementioned issues, we introduce UniPortrait, a solution for unified single- and multi-human image personalization with high-fidelity identity preservation, substantial facial editability, free-form input description, and diverse layout generation (see \figref{fig:highlight}). UniPortrait consists of only two plug-and-play modules: an ID embedding module and an ID routing module. The ID embedding module extracts versatile high-fidelity facial features for each ID and embeds them into the context space of diffusion models. The ID routing module then combines and distributes these embeddings adaptively to their respective regions within the synthesized image, achieving the customization of single and multiple IDs without the prompt and layout restrictions.

More concretely, instead of the last global feature, we utilize the discriminative features with spatial structure from the penultimate layer of the face recognition backbone as the base intrinsic ID features in the ID embedding module. 
To enhance the face fidelity, we further employ CLIP local features following previous works~\cite{ye2023ip-adapter,wu2024infinite} but additionally incorporate shallow features that are empirically low-level and more identity-relevant from the face backbone as the face structure representations. 
Note that contrary to intrinsic ID features, these face structure representations lack the disentanglement and may couple with other identity-irrelevant information such as gaze, pose, or even face-irrelevant features like lighting. To prevent the model from overfitting these spurious facial details, we explicitly decouple the face structure features from the intrinsic ID features by emphasizing a strong dropping regularization, \ie, DropToken~\cite{akbari2021vatt} \& DropPath~\cite{huang2016deep} on the face structure branch. This forced disentangling strategy makes the model more reliant on the intrinsic ID representation, which in turn, affords us increased flexibility in achieving an optimal balance between ID similarity and facial editability. 

In the ID routing module,  to prevent feature mixing between IDs, we propose a routing network to adaptively route and assign a unique ID to each potential face area. 
Specifically, we predict a discrete probability distribution at each spatial location within the cross-attention layer and, during the forward, select the best-matched top-1 ID embedding to engage in the attention of that specific location.
To ensure that all ID conditions are routed and only routed to one target face area, we further introduce a routing regularization loss to assist the router learning. 
Thanks to the adaptiveness of the ID routing module, we allow UniPortrait to customize multi-ID images without any layout predetermination and prompt format restriction, unlocking the creativity of generative models and freedom of prompt designs.

The overall training process of our framework is curated into two stages: the single-ID training stage and the multi-ID fine-tuning stage. The former trains the ID embedding module and the latter specializes in the ID routing module. 
After completing two-stage training, our model can be used for either single-ID customization or multi-ID personalization.
Extensive experiments demonstrate our method's advantages over existing approaches and its good scalability, \eg, the universal compatibility with existing generative control tools such as ControlNet~\cite{zhang2023adding} and IP-Adapter~\cite{ye2023ip-adapter}. 

The contributions of this paper are summarized as below:

\begin{itemize}
	\item We propose UniPortrait, an innovative human image personalization framework that unifies single- and multi-ID customization with high face fidelity and controllability;

	\item We propose a novel ID embedding module with a decoupling strategy, which embeds detailed face identity information while maintaining good editability.

	\item We introduce the ID routing mechanism, which addresses the identity blending issue in multi-ID customization yet without compromising each identity integrity, generated image diversity, and prompt design flexibility.
\end{itemize}

\section{Related Work}

\textbf{Text-to-image generation.}
The development of diffusion models~\cite{dhariwal2021diffusion,ho2020denoising,podell2023sdxl,sohl2015deep,song2019generative} has significantly advanced recent progress in text-conditioned image generation~\cite{chen2023gentron,goodfellow2020generative,kang2023scaling,radford2015unsupervised,ramesh2022hierarchical,ldm,saharia2022photorealistic,sauer2023stylegan,esser2024scaling}. State-of-the-art text-to-image models~\cite{podell2023sdxl,saharia2022photorealistic,betker2023improving} have been able to generate images precisely following user-provided prompts. However, generating images based solely on text cannot produce user-specific concepts, such as their beloved pets or personal items. Consequently, recent research has introduced personalized generation tasks~\cite{textual_inversion,gal2023encoder,kumari2023multi,liu2023cones,ruiz2023dreambooth,wei2023elite}. Among these, identity-preserving human image customization has emerged as a hot and popular topic due to its broad applications.

\noindent\textbf{Single-ID personalization.} 
Early works like FaceStudio~\cite{yan2023facestudio} and InstantID~\cite{wang2024instantid} utilize the global feature of the face backbone as the human ID condition. Due to the loss of spatial information, these methods struggle to capture the intricate details of facial shape and texture. IP-Adapter-FaceID-Plus~\cite{ye2023ip-adapter} and Infinite-ID~\cite{wu2024infinite} introduce local features of CLIP image encoder to enhance face structure representation. Despite their improvements in ID similarity, these methods suffer from subject overfitting due to the insufficient disentanglement between face identity and identity-irrelevant representations. FlashFace~\cite{zhang2024flashface} exploits ReferenceNet~\cite{refonly} but relies on constructing a large dataset containing multiple images of a single ID. CapHuman~\cite{liang2024caphuman} customizes human images based on 3D reconstruction~\cite{li2017learning}, with the render parameters, such as head pose and face position, needing to be set in advance. Most of these methods~\cite{li2024photomaker, peng2024portraitbooth,zhang2024flashface,ye2023ip-adapter,wang2024instantid,yan2023facestudio} focus on single ID customization and encounter difficulties when generalizing to multi-IDs.

\noindent\textbf{Multi-ID personalization.} 
One challenge in customizing multi-ID images is identity blending~\cite{fastcomposer}. A common solution~\cite{avrahami2023break,kumari2022multi,jang2024identity} to this problem involves encoding and integrating all subjects' information into text embeddings, using the text's structure to differentiate different IDs. The drawback of this approach lies in its restriction of prompt formats, \eg, we should explicitly specify the subject word and restrict its representation solely to the singular form. Moreover, such integration can compromise identity fidelity and prompt consistency due to the distinct nature of visual and textual signals. Another line of research~\cite{liu2023cones,kim2024instantfamily,kwon2024concept,kong2024omg,gu2024mix} introduces predefined layout masks to separate the information of different IDs. These methods effectively prevent identity blending but limit the diversity of generated images. FastComposer~\cite{fastcomposer} utilizes localized cross-attention maps to customize multiple IDs without layout constraints. However, it necessitates a specialization of the diffusion model, making it impractical to employ other foundation text-to-image models or control tools from the community. MoA~\cite{ostashev2024moa} draws on the FastComposer principle and deploys a mixed-attention module to disentangle the prior and personalization branches. Despite improving usability, the entanglement of text and image embeddings still results in a less-than-ideal balance between face fidelity and prompt consistency.

Our proposed UniPortrait exploits an independent, versatile ID embedding module with a decoupling strategy to achieve a pleasing trade-off between ID fidelity and prompt consistency,  and meanwhile employs an ID routing module to unify single-ID and multi-ID customization by which we do not specify the ID locations in advance or impose constraints on subject text descriptions.

\section{Methods}

\begin{figure*}[t]
	\centering
	\includegraphics[width=0.95\linewidth]{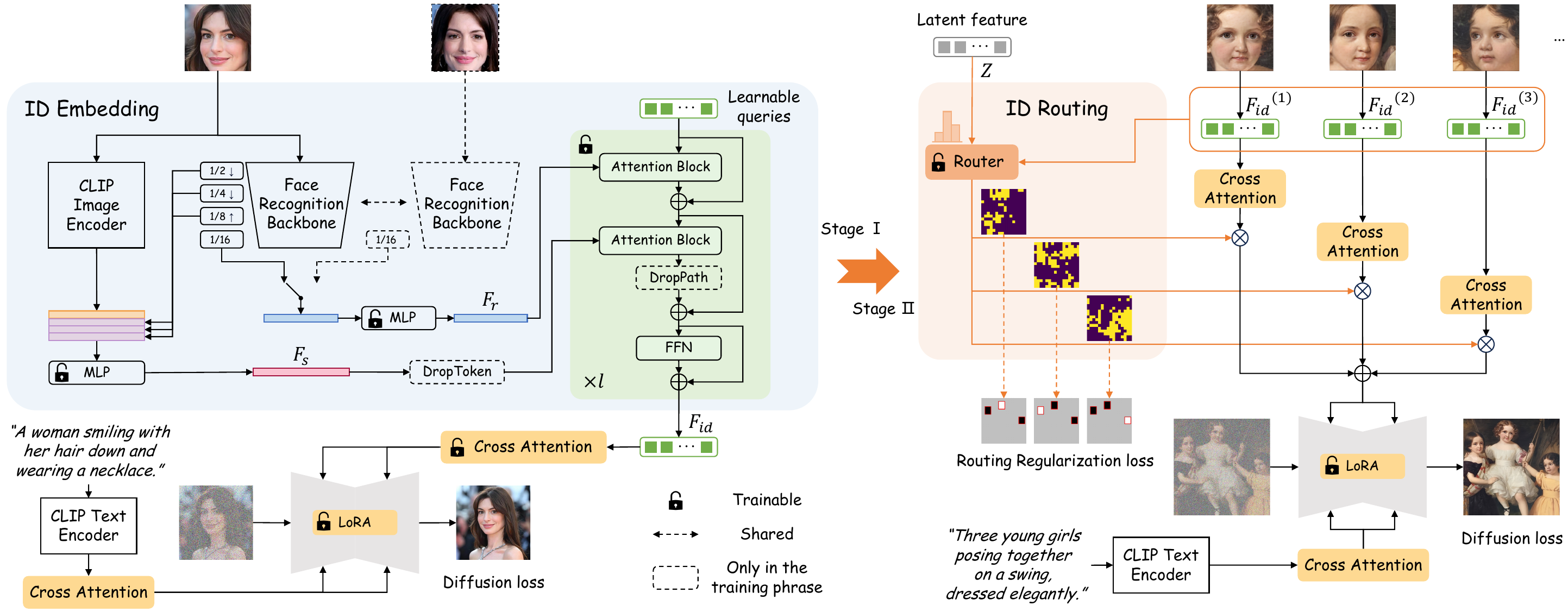}
	\caption{
		\textbf{Overview of UniPortrait framework}. Our proposed UniPortrait consists of two plug-and-play modules: an ID embedding module and an ID routing module. The ID embedding module extracts versatile editable facial features with a decoupling strategy for each ID (\secref{sec:3.2}), and the ID routing module combines and distributes these embeddings to their respective locations adaptively without the intervention for prompts and layouts (\secref{sec:3.3}). The entire training process of the framework is curated into two stages, \ie, the single-ID training stage and the multi-ID fine-tuning stage (\secref{sec:3.4}).
		}
	\label{fig:framework}
\end{figure*}

This section introduces UniPortrait, an innovative approach to unified single-ID and multi-ID character generation. We first briefly review the background of Stable Diffusion in \secref{sec:3.1} and then elaborate on the details of two key modules of UniPortrait in \secref{sec:3.2} and \secref{sec:3.3}. Finally, we illustrate the training scheme of UniPortrait in \secref{sec:3.4}.  The overview of our framework is shown in \figref{fig:framework}.

\subsection{Preliminary}
\label{sec:3.1}

In this paper, the underlying model used for text-to-image synthesis is Stable Diffusion~\cite{ldm}. It takes a text prompt $P$ as input and generates the corresponding image $x_0$. Stable diffusion comprises three main modules: an autoencoder $(\mathcal{E}(\cdot), \mathcal{D}(\cdot))$, a CLIP text encoder $\tau(\cdot)$, and a U-Net $\epsilon_{\theta}(\cdot)$. Typically, it is trained under the following diffusion loss:
\begin{equation}
	\mathcal{L}_{diff} = \mathbb{E}_{z_0,P,\epsilon \sim \mathcal{N}(0,1),t}\lbrack\Vert\epsilon-\epsilon_\theta (z_t,t,\tau(P))\Vert_2^2\rbrack
	\label{eq:1}
\end{equation}
where $\epsilon \sim \mathcal{N}(0,1)$ is the randomly sampled Gaussian noise, $t$ is the time step,  $z_0=\mathcal{E}(x_0)$ is the latent representation of $x_0$, and $z_t$ is calculated by $z_t=\alpha_t z_0+\sigma_t\epsilon$ with the coefficients $\alpha_t$ and $\sigma_t$ provided by the noise scheduler.

\subsection{ID embedding}
\label{sec:3.2}

The ID embedding module is specifically crafted to impart high-fidelity editable face ID information, thereby guiding the diffusion model to generate ID-consistent and -controllable images. Unlike most preceding approaches~\cite{yan2023facestudio, wang2024instantid} that harness the final global features of a face recognition backbone for face ID representation, we utilize features from the penultimate layer (prior to the fully connected layer). This adjustment aims to preserve an enhanced degree of spatial information pertaining to ID features.
Since the face recognition backbone~\cite{deng2019arcface,huang2020curricularface,wang2018cosface,liu2017sphereface} is commonly trained on a large dataset that contains millions of human IDs, its features are expected to be insensitive to ID-irrelevant facial information such as expression, pose, and gaze. In particular, they are insensitive to facial shape and texture details. This is achieved by only noting that the fluctuations in weight or age influence one's appearance but do not change his or her identity. We term these recognition features the intrinsic ID features.

However, there often arises a user demand to personalize ID images to closely match the appearance of a given face reference, \ie, maintaining consistent facial shape and texture details beyond the intrinsic ID characteristics. 
In response, several prior studies~\cite{ye2023ip-adapter,wu2024infinite,cui2024idadapter} have leveraged local features derived from the CLIP image encoder as face structure conditions. Despite the enhanced face similarity, utilizing CLIP local features poses two significant challenges. Firstly, CLIP is trained on generically weakly aligned image-text pairs, rendering its features less discriminative concerning face identities and predominantly semantic. Secondly, due to the lack of disentanglement, these features may couple with other ID-irrelevant facial information or even face-irrelevant representations like background lighting. 
Given the typically scarce and non-diverse nature of personalization training data—in which the training reference and target faces often come from the same or similar images—these irrelevant features risk leading to model overfitting on non-essential facial details, which, in turn, complicates the process of facial control and editing.

In order to solve these problems, we initially integrate the shallow features from the face recognition model to augment the structural representation of the face. Subsequently, we apply a strong dropping regularization to the structure feature branch to decouple it from the intrinsic ID branch. The shallow features of the backbone are empirically low-level, containing more texture details, and they are more ID-relevant, facilitating us to generate higher-fidelity portraits. The dropping regularization on the facial structure branch maintains the independence of intrinsic ID features and face structure features, simultaneously making the model more reliant on intrinsic ID characteristics. Such a strategy allows for a more versatile trade-off between ID similarity and editability, catering to the varied requirements of users seeking identity-preserving portrait generation.

Specifically, as depicted in the blue section of \figref{fig:framework}, we first flatten and apply a Multilayer Perceptron (MLP) to the penultimate layer's features of the face recognition model to obtain the intrinsic ID features $F_r\in\mathbb{R}^{m_r\times d_r}$, where $m_r$ represents the feature length and $d_r$ represents the feature dimension. We then interpolate the shallow features, \ie, the $1/2$, $1/4$, and $1/8$ feature maps from the face backbone and concatenate them with CLIP local features to derive the face structure features $F_s\in\mathbb{R}^{m_s\times d_s}$ through another MLP. Next, we introduce a $l$ layer Q-Former~\cite{li2023blip,li2024blipdiffusion} with $m$ learnable queries to aggregate $F_r$ and $F_s$. Each layer of the Q-Former comprises two attention blocks and one Feed-Forward Network (FFN), with the attention blocks respectively attending to the intrinsic ID information and face structure representations.  In the input and output of the second attention block, we further introduce DropToken~\cite{akbari2021vatt} and DropPath~\cite{huang2016deep} as means of decoupling face structure from intrinsic ID representation. The final output from the Q-Former, denoted as $F_{id}\in\mathbb{R}^{m\times d}$,  is then employed as the ID embedding and aligned into the context space of U-Net. Here, we use decoupled cross attention~\cite{ye2023ip-adapter} to inject the ID information into U-Net.

\noindent\textbf{Single-ID multi-reference embedding.} Owing to the scalability of the designed architecture, our ID embedding module can be seamlessly adapted to cater to the context of embedding multiple references for a single ID. We only need to extract the intrinsic ID features ${F_r}_{(j)}$ and face structure features ${F_s}_{(j)}$  for each reference ($j$ is the reference index) and concatenate them as the new intrinsic ID and face structure features. Experiments show that although our method is trained on a single reference image, it can be effectively extended to accommodate multiple reference images and achieve improved personalization results. See experiments in \appref{app:B} for details.

\noindent\textbf{Identity interpolation.} We extract the ID embedding ${F_{id}}^{(n)}$  for each identity ($n$ is the identity index) and perform linear interpolation on them to achieve ID interpolation. By identity interpolation, we can create a meaningful semantic transition between different IDs or even different states of the same ID. See the experiment in \secref{sec:4.4} for details.

\subsection{ID Routing}
\label{sec:3.3}

Through the ID embedding module, we can obtain versatile editable embeddings for a single ID. For multi-ID scenarios, we leverage the ID embedding module to embed each ID information into the context space. Notably, these embedded ID representations are position-independent, as we have not imposed any positional constraints on them. To avoid identity blending, previous methods have either integrated ID embeddings into text embeddings~\cite{avrahami2023break,kumari2022multi,jang2024identity} or employed manually crafted layout masks~\cite{liu2023cones,kim2024instantfamily,kwon2024concept} to segregate the information of different IDs. The former requires adherence to a specific format of text descriptions (\eg, the singular phrase for the subject) and may potentially degrade the fidelity of both textual and identity representations; the latter constrains the diversity of the resultant imagery. In this work, we introduce a position-wise ID routing module integrated within each cross-attention layer to adaptively route and assign a unique ID to each potential face area in the latent features, thereby effectively mitigating the problem of identity blending.

Specifically, assume there are $N$ distinct IDs, with each ID embedding denoted as ${F_{id}}^{(n)}$, $n=1, \cdots, N$. For every spatial position $(u,v)$ within the latent feature $Z\in \mathbb{R}^{c\times h\times w}$, we route and assign a unique identity $k^*$ ($1\le k^*\le N$) to it with $k^*$ determined as:
\begin{equation}
	 k^*=\mathop{\arg\max}\limits_{k} {\psi (Z,{F_{id}}^{(:)},(u,v))}_k,
	 \label{eq:2}
\end{equation}
where $\psi$ represents the router and it outputs a $N$-dimensional discrete probability distribution as:
\begin{equation}
	\begin{split}
\psi(Z,{F_{id}}^{(:)},&(u,v)) =\\
\operatorname{Softmax}&(\lbrack{\theta(Z)}_{u, v}*\phi(W_{aggr}*{F_{id}}^{(n)})\rbrack_{n=1}^N).
	\end{split}
	\label{eq:3}
\end{equation}
Here $W_{aggr}\in\mathbb{R}^m$ is a weight that aggregates $m$ ID embedding tokens into a singular token, $\theta$ and $\phi$ are two small networks, and $*$ is the matrix multiplication operator. Subsequently, the ID information, as pinpointed by each spatial location, is integrated into the latent feature of that location via the cross-attention mechanism, analogous to the single ID information injection outlined in \secref{sec:3.2}. In practice, we instantiate $\theta$ and $\phi$  as two 2-layer MLPs for simplicity. 

The idea behind the ID routing is that each face in an image is associated with at most one ID feature. By confining each position to cross-attend to solely one ID information, 
the blending problem between IDs is efficaciously circumvented. 

However, directly applying \equref{eq:2} presents two potential concerns. Primarily, it does not guarantee that all IDs will be routed. Secondly, the same ID still has the possibility of being leaked to multiple target faces by attending to their partial areas. Additionally, the \equref{eq:2} is non-differentiable. 
In order to alleviate these issues, we propose the incorporation of a routing regularization loss and leverage the Gumbel softmax trick~\cite{jang2016categorical} during the training phase. These measures facilitate router learning, enhancing its capability to effectively manage and distribute ID representations.

\noindent\textbf{Routing regularization loss.} Concretely, given a target image that contains $N$ distinct IDs during the training phase, we first detect bounding boxes for all faces on the image and convert them into binary masks, where 1 represents the face area and 0 represents the other area. In this way, we obtain $N$ face region masks. Then, the routing regularization loss is calculated by the L2 loss between the router's outputs and these face region masks as follows:
\begin{equation}
\mathcal{L}_{route}=\lambda\cdot\frac{1}{N}\Vert W_{route}\odot(\psi(Z,{F_{id}}^{(:)}) -M)\Vert^2_2
\label{eq:4}
\end{equation}
where $\lambda$ is the weight of the loss function,  $\odot$ denotes the element-wise multiplication, $\psi(Z,{F_{id}}^{(:)})\in\mathbb{R}^{N \times h\times w}$ represents the routing outputs across all positions, $M\in\mathbb{R}^{N\times h\times w}$ is the stack of $N$ face region masks, and $W_{route }\in\mathbb{R}^{h\times w}$ is the union of $N$ face region masks which means that we only apply regularization loss to the face regions in the image.
By emphasizing the routing targets across all face region areas, on the one hand, we encourage all IDs to be routed, and on the other hand, we prompt the ID router that each ID can merely be routed to at most one face area.

\noindent\textbf{Gumbel softmax trick.} To ensure the gradients of the routing module can be properly backpropagated during training, we introduce the Gumbel-softmax trick~\cite{jang2016categorical}. Specifically, during training, we add Gumbel noise to the output logits of the router to reparameterize the routing sampling process. During inference, we normally select the best-determined top-1 identity from the router for forward propagation.

Finally, it is noteworthy that in the case of a single ID, the router becomes trivial, and the routing-based multi-ID generation degenerates into common single-ID generation.

\begin{figure*}[t]
	\centering
	\includegraphics[width=1.0\linewidth]{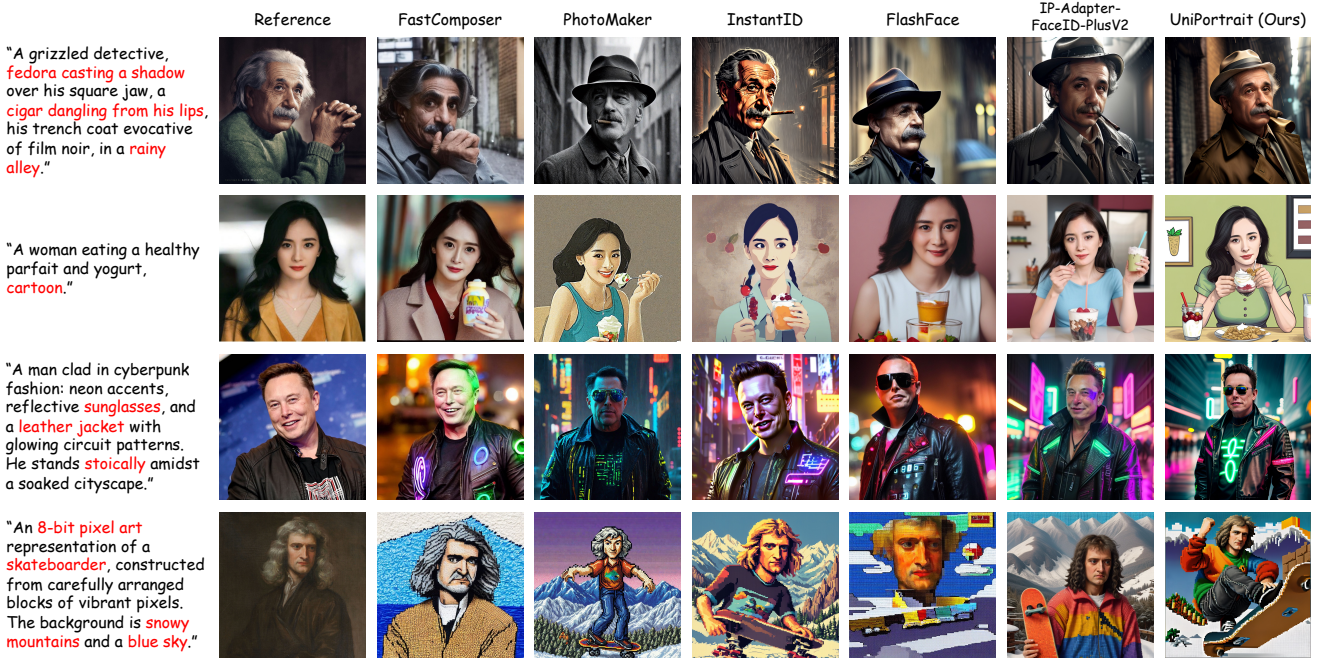}
	\caption{\textbf{Qualitative comparison of different methods on single-ID image customization.}}
	\label{fig:single_qualitative}
\end{figure*}

\begin{table}[t]
	\centering
	\tablestyle{4pt}{1.2}\scriptsize\begin{tabular}{y{55} | x{15} x{14} x{28}  x{22}  x{20}  x{22}}
		\multirow{2}*{Method} & \multirow{2}*{Arch. }  & \multirow{2}*{\shortstack{Multi-\\ ID?}} & \multirow{2}*{\shortstack{Face Sim. \\$\uparrow$ (\%) }} & \multirow{2}*{\shortstack{CLIP-T \\ $\uparrow$ (\%)}} & \multirow{2}*{FID $\downarrow$} & \multirow{2}*{\shortstack{LAION-\\Aes $\uparrow$}} \\
		&&&&&&\\
		\shline
		\demph{SD v1-5~\cite{ldm}} & 	\demph{SD15} & 	\demph{-} & 	\demph{\phantom{0}4.4} & 	\demph{27.7} &	\demph{ -} & 	\demph{6.60} \\
		PortraitBooth$^\dagger$~\cite{peng2024portraitbooth} & SD15 & \XSolidBrush & 65.7	& 24.5 & - & - \\
		IP-Adapter$^*$~\cite{ye2023ip-adapter} & SD15 & \XSolidBrush & 68.4 & 24.7 & 139.5 & \textbf{6.43} \\
		FlashFace~\cite{zhang2024flashface} & SD15 & \XSolidBrush & \underline{72.7} & 25.8 & 141.7 & 5.85  \\
		FastComposer~\cite{fastcomposer} & SD15 & \Checkmark & 50.8 & 24.1 & \underline{134.5} & 6.17 \\	
		\gr
		\textbf{\modelname} (ours) & SD15 & \Checkmark & 71.1 & \underline{26.1} & \textbf{123.4} & \underline{6.42}  \\
		PhotoMaker~\cite{li2024photomaker} & SDXL & \XSolidBrush & 41.7 & \textbf{26.9} & 136.1 & 6.01 \\
		InstantID~\cite{wang2024instantid} & SDXL & \XSolidBrush & \textbf{77.9} & 24.1 & 163.5 & 6.15  \\
	\end{tabular}
	\vspace{-1mm}
	
	\caption{\textbf{Quantitative comparison with different single-ID customization methods.} $^\dagger$ denotes the results reported in its original paper. IP-Adapter$^*$ corresponds to the IP-Adapter-FaceID-PlusV2 variant. ``Multi-ID'' indicates the capability of the method to customize multi-ID images using only text and ID conditions. The best result is shown in \textbf{bold}, and the second best is \underline{underlined}.}
	
	\vspace{-3mm}
	
	\label{tab:single-id-quatitative}
\end{table}

\subsection{Training}
\label{sec:3.4}

The entire training process of UniPortrait is curated into two stages: the single-ID training stage and the multi-ID fine-tuning stage. After completing this two-stage training, UniPortrait can be used for either single-ID customization or multi-ID personalization.

\noindent\textbf{Stage \uppercase\expandafter{\romannumeral1}: single ID training.} In this phase, we only introduce the ID embedding module; the training regimen is limited to images that feature a singular ID, as is depicted on the left side of \figref{fig:framework}. We first crop and align the face region of an image to serve as the input for the ID embedding module. If the face has an associated ID label, \eg, an image sourced from the CelebA dataset~\cite{liu2015deep}, another cropped and aligned face image of the identical ID is sampled with a 0.1 probability to act as the input for the intrinsic ID branch. Conversely, all inputs for the face structure branch are harnessed from the target image, an approach adopted to enhance learning of the textural and structural details of the face.  Throughout the training process, we employ a dropping regularization on the face structure branch, with probabilities delineated as follows: complete branch dropout occurs with a 0.33 probability; retaining the branch while randomly dropping face structure tokens at a 0.33 probability; and complete preservation of the face structure branch at a 0.34 probability. To extract facial information more comprehensively, Low-Rank Adaptation (LoRA~\cite{hu2022lora}) has been appended to the U-Net architecture.  Only the parameters within the ID embedding module and the U-Net's LoRA are subjected to training in this stage. The training loss is aligned with the conventional diffusion loss, as shown in \equref{eq:1}.

\noindent\textbf{Stage \uppercase\expandafter{\romannumeral2}: multiple ID fine-tuning.} After completing the Stage \uppercase\expandafter{\romannumeral1} training, we introduce the ID routing module. We fix all the parameters in the ID embedding module and only fine-tune the parameters of the ID router and LoRA module, with the learning rate of LoRA module decaying by 0.1. The loss function in the second stage encompasses the original diffusion loss (\equref{eq:1}) and the routing regularization loss (\equref{eq:4}). Herein, the balancing parameter $\lambda$ is set to 0.1.

\begin{figure*}[t]
	\centering
	\includegraphics[width=1.0\linewidth]{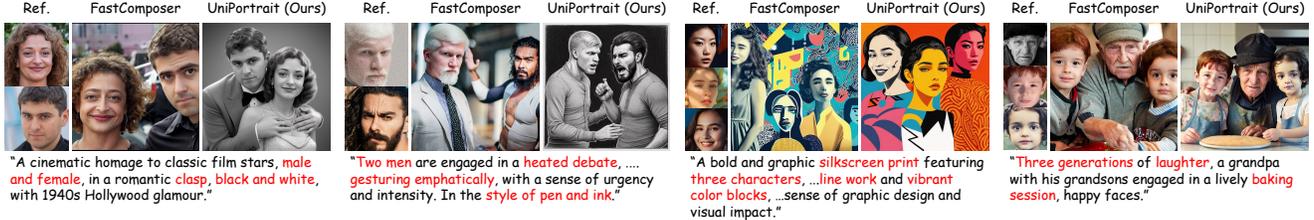}
	\vspace{-8mm}
	\caption{\textbf{Qualitative comparison of different methods on multi-ID image customization.} For compatibility with FastComposer, numerical plural expressions (\eg, ``two men'') are converted into singular phrases linked by ``and'' (\eg, ``a man and a man'').}
	 \vspace{-3mm}
	\label{fig:multiple_qualitative}
\end{figure*}

\section{Experiments}

\subsection{Setup}

\noindent\textbf{Dataset.} The dataset utilized in this work comprises four primary segments: (1) 240k single-ID images filtered from LAION~\cite{schuhmann2022laion,schuhmann2021laion}; (2) 100k single-ID portraits filtered from CelebA database~\cite{liu2015deep}; (3) 160k high-quality single-ID images collected from the Internet; (4) 120k high-quality multi-ID portraits filtered from LAION. The first three subsets are utilized for Stage \uppercase\expandafter{\romannumeral1} training, while the last subset is used for Stage \uppercase\expandafter{\romannumeral2}  training. Images originating from CeleA and those obtained from the Internet are captioned using Qwen-VL~\cite{bai2023qwen}, whereas images from LAION are presented with their original text captions.  It is noteworthy that within all these data, only the CeleA images are endowed with ID annotations.

\begin{figure}[t]
	\centering
	\includegraphics[width=1.0\linewidth]{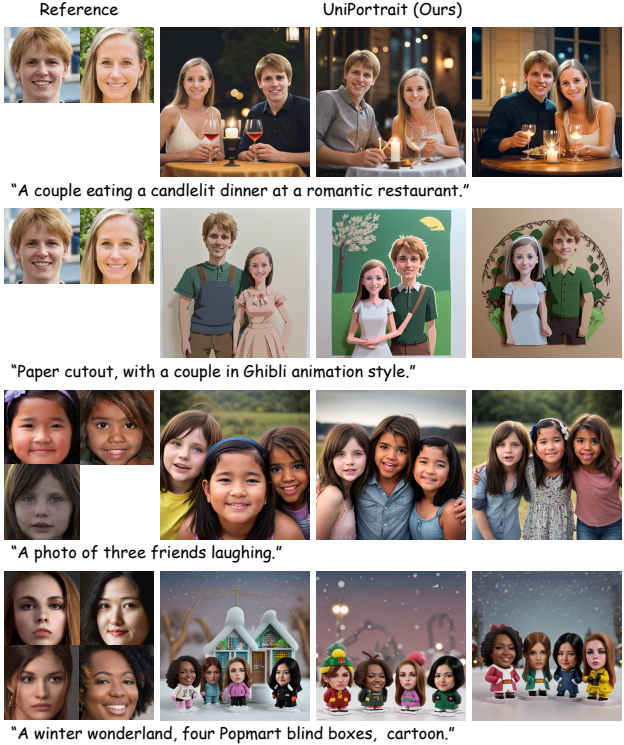}
	\vspace{-6mm}
	\caption{\textbf{Additional examples of multi-ID customization.} UniPortrait is capable of customizing multi-ID images using free-form prompts and generating diverse layouts.}
	\vspace{-3mm}
	\label{fig:multiple_qualitative_plus}
\end{figure}

\noindent\textbf{Implementation details.} Our training starts with the StableDiffusion v1-5~\cite{ldm} model. The face recognition backbone we used is CurricularFace~\cite{huang2020curricularface}. For CLIP image encoder, we use OpenCLIP's \texttt{clip-vit-huge-patch14}. The Q-Former in ID embedding module has 6 layers and 16 leanable queries. The rank of LoRA in U-Net is set to 128. All experiments are conducted on 8 V100 GPUs using AdamW~\cite{loshchilov2017decoupled} optimizer with a batch size of 128 and a learning rate of 1e-5. The first stage trains 300k iterations, and the second stage trains 150k iterations. To facilitate classifier-free guidance sampling~\cite{ho2022classifier}, we train the model without the face conditions on 5\% images. During inference, we use 20-step DDIM~\cite{song2020denoising} sampling with a classifier-free guidance scale of 7.5, and to achieve more realistic image generation, we employ the \textit{Realistic Vision V4.0} model from \textit{huggingface} following the previous work~\cite{ye2023ip-adapter}.

\noindent\textbf{Evaluation Metrics.} We assess the image generation quality in terms of identity preservation, prompt consistency, FID~\cite{heusel2017gans}, and LAION-Aesthetics (LAION-Aes) scores~\cite{LAION-Aesthetics}. For identity preservation and prompt consistency, we follow the evaluation protocol established by FastComposer~\cite{fastcomposer}. Specifically, identity preservation is quantified by calculating pairwise face similarities between reference and generated faces with FaceNet~\cite{schroff2015facenet}. For multi-identity generation, we detect all faces in generated images and use a greedy matching procedure between generated and reference faces. The lowest similarity score across all faces measures overall identity preservation. Prompt consistency is assessed utilizing the average CLIP-L/14 image-text similarity.

\begin{table}[t]
	\centering
	\tablestyle{4pt}{1.2}\scriptsize\begin{tabular}{y{59} | x{19} x{18} x{32}  x{16}  x{24}}
		\multirow{2}*{Method} & \multirow{2}*{Arch. }  &  \multirow{2}*{\shortstack{Face Sim. \\$\uparrow$ (\%) }} & \multirow{2}*{\shortstack{CLIP-T \\ $\uparrow$ (\%)}} & \multirow{2}*{FID $\downarrow$} & \multirow{2}*{\shortstack{LAION-\\Aes $\uparrow$}} \\
		&&&&&\\
		\shline
		\demph{SD v1-5~\cite{ldm}} & \demph{SD15} &  \demph{1.6} & \demph{28.9} & \demph{-} & \demph{6.25} \\
		FastComposer~\cite{fastcomposer} & SD15  & 38.0 & 25.5 & 156.6 & 5.81 \\	
		\gr
		\textbf{\modelname} (ours) & SD15  & \textbf{67.3} & \textbf{27.4} & \textbf{139.5} & \textbf{5.89}  \\
	\end{tabular}
	\vspace{-1mm}
	
	\caption{\textbf{Quantitative comparison with different multi-ID customization methods.}}
	
	\vspace{-3mm}
	
	\label{tab:multi-id-quatitative}
\end{table}

\begin{table*}[t]
	\centering
	\tablestyle{4pt}{1.2}\scriptsize\begin{tabular}{x{40} x{40} | x{40}x{40}x{40}| x{50} x{50}  x{30}  x{50}}
		
		\multicolumn{2}{c|}{Intrisic ID branch} & \multicolumn{3}{c|}{Face structure branch}
		&  \multirow{3}*{ Face  Sim. $\uparrow$ (\%) } & \multirow{3}*{ CLIP-T $\uparrow$ (\%) } & \multirow{3}*{FID $\downarrow$} & \multirow{3}*{LAION-Aes $\uparrow$} \\

		\multirow{2}*{\shortstack{face backbone \\ global feat.}}  &  \multirow{2}*{\shortstack{face backbone \\ local feat.}} & \multirow{2}*{\shortstack{CLIP \\ local feat}}  & 
		\multirow{2}*{\shortstack{face backbone \\ shallow feat.}} &  \multirow{2}*{\shortstack{drop token \\ \& path}} &&&&\\
		&&&&&&&&\\
		\shline
		\Checkmark &                        &                       &                        &                       &  45.5 &  \textbf{27.2} &  \textbf{115.2} & 6.47 \\
		& \Checkmark &                       &                        &                       & 58.3 & 26.8 & 120.3 &  \textbf{6.50} \\	
		& \Checkmark & \Checkmark &                       &                        &  61.4 & 26.1 & 124.7 & 6.38 \\
		& \Checkmark & \Checkmark & \Checkmark &                        &  64.5 & 25.9 & 128.1 & 6.46 \\
		& \Checkmark & \Checkmark &                        & \Checkmark &  68.4 & 26.1 & 122.6 & 6.30 \\
		& \Checkmark & \Checkmark & \Checkmark & \Checkmark &   \textbf{71.1} & 26.1 & 123.4 & 6.42 \\
	\end{tabular}
	\vspace{-1mm}
	
	\caption{\textbf{Ablation studies for components in ID embedding module.} }
	
	\vspace{-3mm}
	
	\label{tab:single-id-ablation}
\end{table*}

\subsection{Results}
\label{sec:4.2}

\noindent\textbf{Single-ID personalization.} We first evaluate the performance of single-ID customization. We follow FastComposer~\cite{fastcomposer} and use 15 identities from the CelebA dataset~\cite{liu2015deep}, which have been deliberately excluded from our training dataset, with 40 unique text prompts assigned to each subject for assessment. These text prompts cover a wide range of scenarios, such as re-contextualization, stylization, accessorization, and various actions. For a fair comparison, all methods accept a single reference face image and generate 4 images each once. The quantitative results are shown in \tabref{tab:single-id-quatitative}. 
Our approach manifests a commendable balance between identity preservation and prompt consistency, simultaneously achieving the lowest FID score and the second-highest score in LAION-Aesthetics, which markedly surpasses the performance metrics of PortraitBooth~\cite{peng2024portraitbooth}, IP-Adapter-FaceID-PlusV2~\cite{ye2023ip-adapter}, and FastComposer~\cite{fastcomposer}. 
Notably, InstantID~\cite{wang2024instantid} records the highest similarity in face identity; however, its scores for prompt consistency and FID are relatively inferior, a limitation attributed to its requisite for fixed positions of facial landmarks. PhotoMaker~\cite{li2024photomaker} showcases notable results in prompt consistency, albeit with mediocre outcomes in facial similarity. While FlashFace~\cite{zhang2024flashface} manages to accomplish a relative trade-off between face similarity and prompt consistency, the inferior FID and LAION-Aes values indicate its unsatisfactory behavior in the quality and diversity of generated images.
It is crucial to note that among all the evaluated methods, only FastComposer and our approach facilitate the personalized generation of images featuring multiple individuals naively. \figref{fig:single_qualitative} presents a visual comparison of the qualitative outcomes derived from utilizing different methods to respond to a series of prompts for single ID personalization, where the qualitative analysis aligns with the conclusions made from quantitative metrics.

\begin{table}[t]
	\centering
	\tablestyle{4pt}{1.2}\scriptsize\begin{tabular}{y{86} | x{18} x{32}  x{16}  x{24}}
		\multirow{2}*{} &  \multirow{2}*{\shortstack{Face Sim. \\$\uparrow$ (\%) }} & \multirow{2}*{\shortstack{CLIP-T \\ $\uparrow$ (\%)}} & \multirow{2}*{FID $\downarrow$} & \multirow{2}*{\shortstack{LAION-\\Aes $\uparrow$}} \\
		&&&&\\
		\shline
		w/o routing regularization loss  &  59.1 & \textbf{27.5} & 139.8 & 5.89 \\
		w/ routing regularization loss & \textbf{67.3} & 27.4 & \textbf{139.5} & \textbf{5.89} \\	
	\end{tabular}
	\vspace{-1mm}
	
	\caption{\textbf{Ablation study for the routing regularization loss.} }
	
	\vspace{-5mm}
	
	\label{tab:multi-id-ablation}
\end{table}

\noindent\textbf{Multi-ID personalization.}
We further assess the performance of multi-ID image generation. We also use the test benchmark from FastComposer~\cite{fastcomposer}, which contains the 15 IDs from CelebA database described above and 21 additionally curated test prompts. These 15 IDs are strategically paired, resulting in a total of 105 multi-ID combinations. \tabref{tab:multi-id-quatitative} shows the quantitative comparison between UniPortrait and FastComposer. Our method outperforms FastComposer on all metrics, manifesting enhanced identity preservation and prompt consistency alongside augmented quality and aesthetics in the generated images. A qualitative analysis is shown in \figref{fig:multiple_qualitative}. UniPortrait retains the distinctive attributes of different subjects. Concurrently, UniPortrait exhibits improved fidelity to the text prompts, enabling the direct application of text for the stylized customization of images featuring multiple persons. In addition, thanks to the ID routing mechanism, our approach supports greater flexibility in prompt input. This is particularly advantageous for inputs comprising plural phrases, which, in the case of FastComposer, necessitate conversion into singular phrases interconnected by ``and''. More visual examples illustrating the versatility of our method in rendering multi-ID images are presented in \figref{fig:multiple_qualitative_plus}, further substantiating the qualitative enhancements our approach brings to multi-ID image customization.

\begin{figure}[t]
	\centering
	\includegraphics[width=1.0\linewidth]{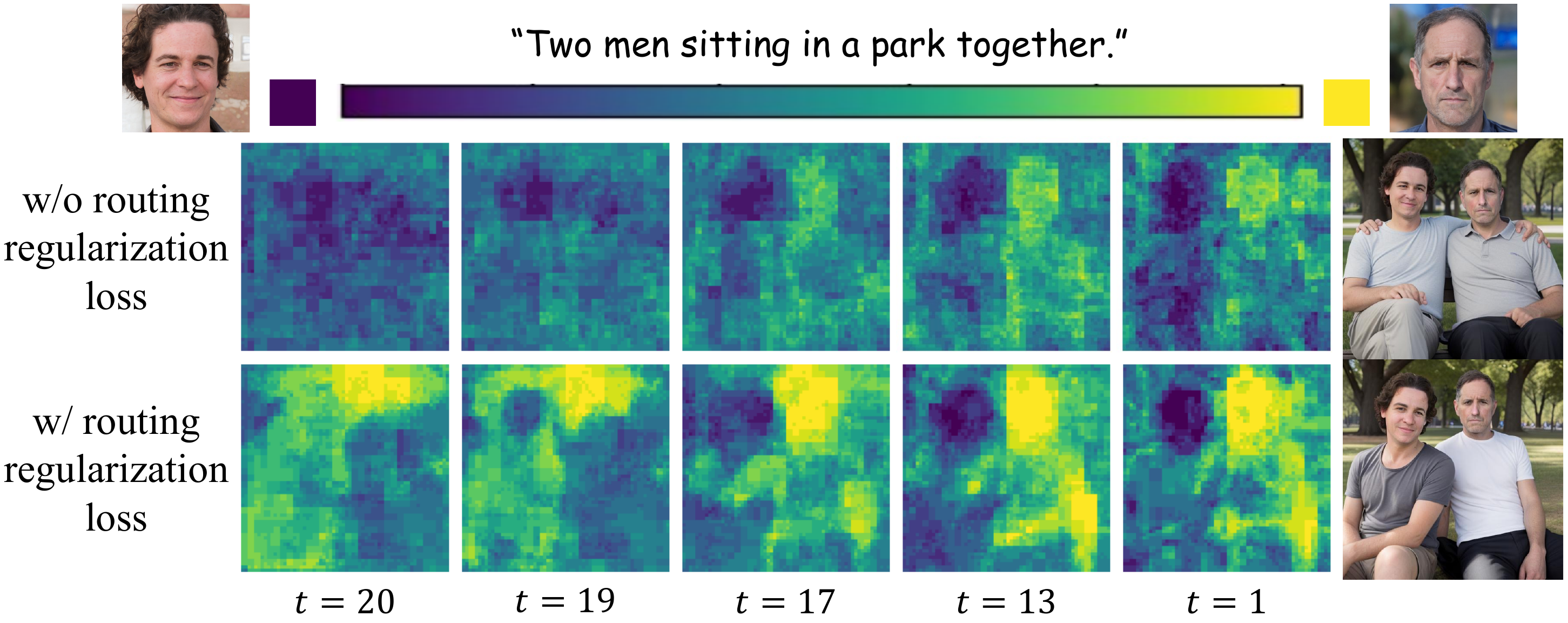}
	\vspace{-6mm}
	\caption{\textbf{The effect of the routing regularization loss.}}
	\vspace{-3mm}
	\label{fig:routing_map}
\end{figure}

\subsection{Ablation study}

\noindent\textbf{Components in ID embedding module.} \tabref{tab:single-id-ablation} presents an assessment of the efficacy of the various constituents within the ID embedding module. Using local features from the penultimate layer of the face recognition model instead of its last global feature contributes to a marked enhancement in ID similarity. Introducing face structure features further bolsters the ID similarity, with this effect appearing particularly pronounced when incorporating shallow features from the face backbone. 
Nevertheless,  it is observed that the integration of facial structure features incurs a decrease in both the diversity of generated images, \ie, FID, and their consistency with the associated text prompts. This reduction can be mitigated by the application of DropToken and DropPath regularizations within the face structure branch. Simultaneously, these regularizations help mitigate model overdependence on inaccurate facial details, thereby optimally reinforcing ID similarity. Despite these adjustments, it must be acknowledged that the inclusion of the face structure branch inherently entails a compromise in prompt consistency to some extent. For more analysis on the branch of face structure, please refer to the \appref{app:A}.

\begin{figure*}[t]
	\centering
	\includegraphics[width=1.0\linewidth]{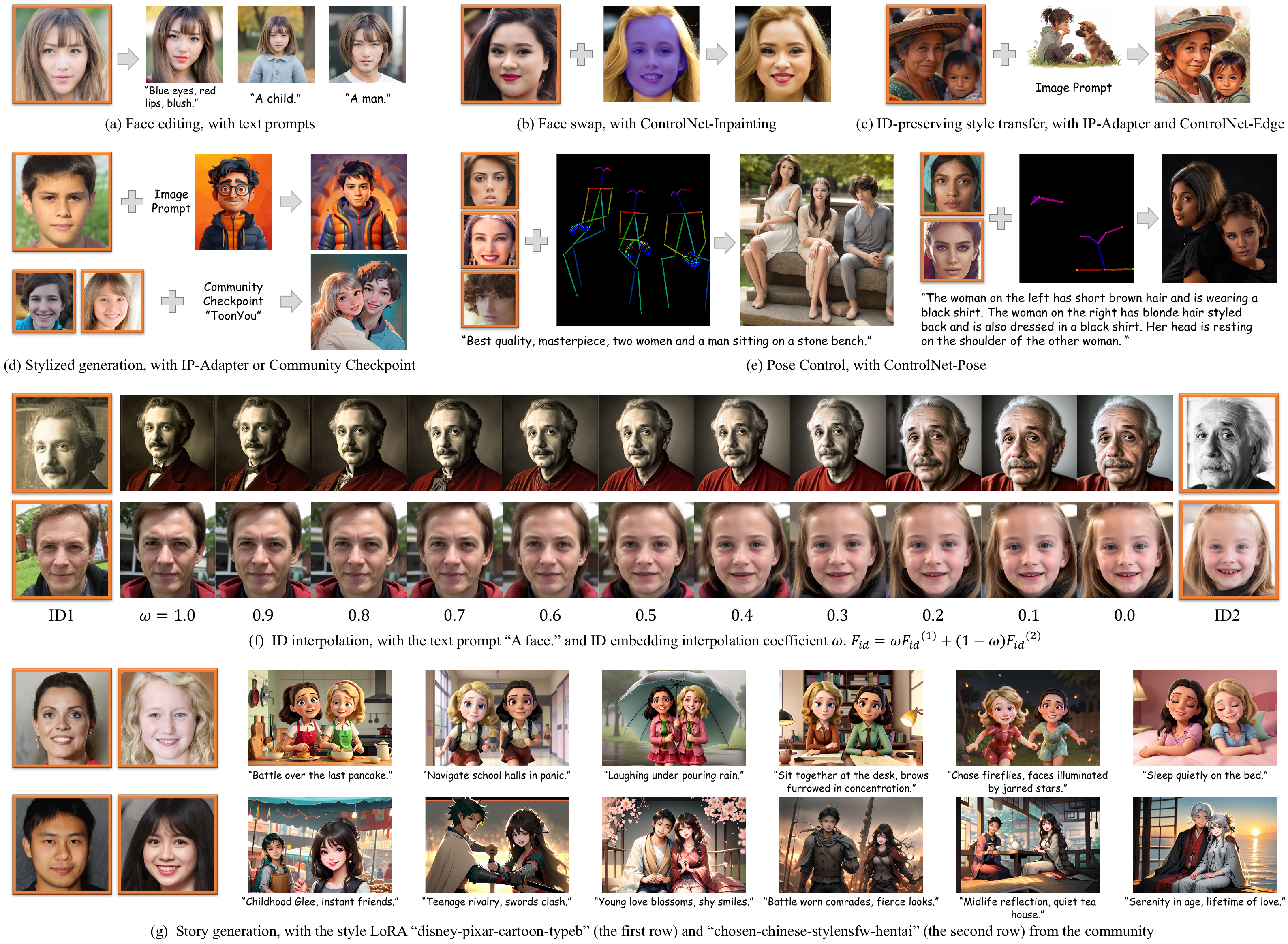}
	\vspace{-6mm}
	\caption{\textbf{Diverse applications of \modelname.}}
	\vspace{-3mm}
	\label{fig:application}
\end{figure*}

\noindent\textbf{Routing regularization loss.}  \tabref{tab:multi-id-ablation} provides a verification of the efficacy of the routing regularization loss. The results suggest that this approach can substantially enhance ID similarity while concurrently maintaining prompt consistency in multi-ID customization. \figref{fig:routing_map} depicts the average routing maps derived from all U-Net cross-attention layers at different diffusion steps. It is observable that the employment of routing regularization loss results in a more focused routing result, indicative of improved segregation of information pertaining to distinct IDs. A more detailed layer-by-layer routing graphs are available for review in the \appref{app:C}.

\subsection{Application}
\label{sec:4.4}
The superior performance of UniPortrait in aligning IDs, maintaining prompt consistency, as well as enhancing the diversity and quality of generated images, paves the way for a plethora of potential downstream applications. Among these, face attribute modification stands out – this includes alterations in age, gender, and specific facial characteristics, as shown in \figref{fig:application}(a).
Additionally, the adaptable, plug-and-play nature of UniPortrait ensures compatibility with a range of existing community-developed tools, such as ControlNet~\cite{zhang2023adding}, LoRA~\cite{hu2022lora}, and IP-Adapter~\cite{ye2023ip-adapter}. This integration facilitates the creation of conditionally controllable ID-preserving generations. Examples of such applications are shown in \figref{fig:application}(b-e).
Going a step further, UniPortrait's capacity for identity interpolation among different characters is investigated, showcasing its adeptness in smoothly blending features from multiple identities, as shown in \figref{fig:application}(f). 
Besides these, we also demonstrate the potential of UniPortrait to generate stories with consistent IDs, as exemplified in  \figref{fig:application}(g).
For more applications and examples, please refer to the \appref{app:D}.

\section{Conclusion}

We introduce UniPortrait, a model developed for the unified customization of single- and multi-ID images. UniPortrait incorporates an advanced ID embedding module that ensures high-fidelity and editable identity embeddings. Furthermore, a modular, plug-and-play ID routing component has been integrated to address the challenge of identity blendings during the multi-ID generation process.
The empirical results demonstrate that UniPortrait outperforms existing methods by delivering a synthesis that is not only of high quality and diversity but also offers robust editability and strong identity fidelity.
We hope that our UniPortrait will serve as a new baseline work within this domain, providing a benchmark that can be followed, replicated, and optimized by all research institutions.

\noindent\textbf{Limitations.} 
Considering that the routing decisions of the router are exclusively based on human ID information, our ID routing module is currently incapable of customizing attributes unrelated to face identity, \eg, clothing and actions, for each ID in the multi-ID generation.
A possible solution is to feed the representations of all attributes of interest into the router to guide the ID routing, \ie, the attribute-binding ID routing. We leave it for future research.

~

\noindent\textbf{Ethics considerations.} The utilization of human image personalization technologies presents potential social risks, with the misuse of such technologies for deepfakes being a significant concern. In order to address these risks, the implementation of ethical guidelines and responsible usage practices is imperative. At present, the synthesized outcomes display certain visual anomalies that may facilitate the identification of deepfakes.

\section*{Acknowledgments}
Most of the face images used in our experiments come from the Pexels, Unsplash, Pixabay, and Wikipedia websites. We thank the owners of these images for sharing their valuable assets. We also thank the StyleGAN2 authors for sharing their high-quality synthesized face images, which constitute another important part of the ID images we used.

{
    \small
    \bibliographystyle{ieeenat_fullname}
    \bibliography{main}

\begin{thebibliography}{70}
\providecommand{\natexlab}[1]{#1}
\providecommand{\url}[1]{\texttt{#1}}
\expandafter\ifx\csname urlstyle\endcsname\relax
  \providecommand{\doi}[1]{doi: #1}\else
  \providecommand{\doi}{doi: \begingroup \urlstyle{rm}\Url}\fi

\bibitem[Akbari et~al.(2021)Akbari, Yuan, Qian, Chuang, Chang, Cui, and
  Gong]{akbari2021vatt}
Hassan Akbari, Liangzhe Yuan, Rui Qian, Wei-Hong Chuang, Shih-Fu Chang, Yin
  Cui, and Boqing Gong.
\newblock Vatt: Transformers for multimodal self-supervised learning from raw
  video, audio and text.
\newblock \emph{NeurIPS}, 34:\penalty0 24206--24221, 2021.

\bibitem[Avrahami et~al.(2023)Avrahami, Aberman, Fried, Cohen-Or, and
  Lischinski]{avrahami2023break}
Omri Avrahami, Kfir Aberman, Ohad Fried, Daniel Cohen-Or, and Dani Lischinski.
\newblock Break-a-scene: Extracting multiple concepts from a single image.
\newblock In \emph{SIGGRAPH Asia 2023 Conference Papers}, pages 1--12, 2023.

\bibitem[Bai et~al.(2023)Bai, Bai, Yang, Wang, Tan, Wang, Lin, Zhou, and
  Zhou]{bai2023qwen}
Jinze Bai, Shuai Bai, Shusheng Yang, Shijie Wang, Sinan Tan, Peng Wang, Junyang
  Lin, Chang Zhou, and Jingren Zhou.
\newblock Qwen-vl: A frontier large vision-language model with versatile
  abilities.
\newblock \emph{arXiv preprint arXiv:2308.12966}, 2023.

\bibitem[Betker et~al.(2023)Betker, Goh, Jing, Brooks, Wang, Li, Ouyang,
  Zhuang, Lee, Guo, et~al.]{betker2023improving}
James Betker, Gabriel Goh, Li Jing, Tim Brooks, Jianfeng Wang, Linjie Li, Long
  Ouyang, Juntang Zhuang, Joyce Lee, Yufei Guo, et~al.
\newblock Improving image generation with better captions.
\newblock \emph{Computer Science. https://cdn. openai. com/papers/dall-e-3.
  pdf}, 2\penalty0 (3):\penalty0 8, 2023.

\bibitem[Chen et~al.(2023)Chen, Xu, Ren, Cong, He, Xie, Sinha, Luo, Xiang, and
  Perez-Rua]{chen2023gentron}
Shoufa Chen, Mengmeng Xu, Jiawei Ren, Yuren Cong, Sen He, Yanping Xie, Animesh
  Sinha, Ping Luo, Tao Xiang, and Juan-Manuel Perez-Rua.
\newblock Gentron: Delving deep into diffusion transformers for image and video
  generation.
\newblock \emph{arXiv preprint arXiv:2312.04557}, 2023.

\bibitem[Cui et~al.(2024)Cui, Guo, An, Deng, Zhao, Wei, and
  Feng]{cui2024idadapter}
Siying Cui, Jia Guo, Xiang An, Jiankang Deng, Yongle Zhao, Xinyu Wei, and
  Ziyong Feng.
\newblock Idadapter: Learning mixed features for tuning-free personalization of
  text-to-image models.
\newblock In \emph{CVPR}, pages 950--959, 2024.

\bibitem[Dahary et~al.(2024)Dahary, Patashnik, Aberman, and
  Cohen-Or]{dahary2024yourself}
Omer Dahary, Or Patashnik, Kfir Aberman, and Daniel Cohen-Or.
\newblock Be yourself: Bounded attention for multi-subject text-to-image
  generation.
\newblock \emph{arXiv preprint arXiv:2403.16990}, 2024.

\bibitem[Deng et~al.(2019)Deng, Guo, Xue, and Zafeiriou]{deng2019arcface}
Jiankang Deng, Jia Guo, Niannan Xue, and Stefanos Zafeiriou.
\newblock Arcface: Additive angular margin loss for deep face recognition.
\newblock In \emph{CVPR}, pages 4690--4699, 2019.

\bibitem[Dhariwal and Nichol(2021)]{dhariwal2021diffusion}
Prafulla Dhariwal and Alexander Nichol.
\newblock Diffusion models beat gans on image synthesis.
\newblock \emph{NeurIPS}, 34:\penalty0 8780--8794, 2021.

\bibitem[Esser et~al.(2024)Esser, Kulal, Blattmann, Entezari, M{\"u}ller,
  Saini, Levi, Lorenz, Sauer, Boesel, et~al.]{esser2024scaling}
Patrick Esser, Sumith Kulal, Andreas Blattmann, Rahim Entezari, Jonas
  M{\"u}ller, Harry Saini, Yam Levi, Dominik Lorenz, Axel Sauer, Frederic
  Boesel, et~al.
\newblock Scaling rectified flow transformers for high-resolution image
  synthesis.
\newblock In \emph{ICML}, 2024.

\bibitem[Gal et~al.(2022)Gal, Alaluf, Atzmon, Patashnik, Bermano, Chechik, and
  Cohen-Or]{gal2022image}
Rinon Gal, Yuval Alaluf, Yuval Atzmon, Or Patashnik, Amit~H Bermano, Gal
  Chechik, and Daniel Cohen-Or.
\newblock An image is worth one word: Personalizing text-to-image generation
  using textual inversion.
\newblock \emph{arXiv preprint arXiv:2208.01618}, 2022.

\bibitem[Gal et~al.(2023{\natexlab{a}})Gal, Alaluf, Atzmon, Patashnik, Bermano,
  Chechik, and Cohen-Or]{textual_inversion}
Rinon Gal, Yuval Alaluf, Yuval Atzmon, Or Patashnik, Amit~H Bermano, Gal
  Chechik, and Daniel Cohen-Or.
\newblock An image is worth one word: Personalizing text-to-image generation
  using textual inversion.
\newblock In \emph{ICLR}, 2023{\natexlab{a}}.

\bibitem[Gal et~al.(2023{\natexlab{b}})Gal, Arar, Atzmon, Bermano, Chechik, and
  Cohen-Or]{gal2023encoder}
Rinon Gal, Moab Arar, Yuval Atzmon, Amit~H Bermano, Gal Chechik, and Daniel
  Cohen-Or.
\newblock Encoder-based domain tuning for fast personalization of text-to-image
  models.
\newblock \emph{ACM TOG}, 42\penalty0 (4):\penalty0 1--13, 2023{\natexlab{b}}.

\bibitem[Goodfellow et~al.(2020)Goodfellow, Pouget-Abadie, Mirza, Xu,
  Warde-Farley, Ozair, Courville, and Bengio]{goodfellow2020generative}
Ian Goodfellow, Jean Pouget-Abadie, Mehdi Mirza, Bing Xu, David Warde-Farley,
  Sherjil Ozair, Aaron Courville, and Yoshua Bengio.
\newblock Generative adversarial networks.
\newblock \emph{Communications of the ACM}, 63\penalty0 (11):\penalty0
  139--144, 2020.

\bibitem[Gu et~al.(2024)Gu, Wang, Wu, Shi, Chen, Fan, Xiao, Zhao, Chang, Wu,
  et~al.]{gu2024mix}
Yuchao Gu, Xintao Wang, Jay~Zhangjie Wu, Yujun Shi, Yunpeng Chen, Zihan Fan,
  Wuyou Xiao, Rui Zhao, Shuning Chang, Weijia Wu, et~al.
\newblock Mix-of-show: Decentralized low-rank adaptation for multi-concept
  customization of diffusion models.
\newblock \emph{NeurIPS}, 36, 2024.

\bibitem[Heusel et~al.(2017)Heusel, Ramsauer, Unterthiner, Nessler, and
  Hochreiter]{heusel2017gans}
Martin Heusel, Hubert Ramsauer, Thomas Unterthiner, Bernhard Nessler, and Sepp
  Hochreiter.
\newblock Gans trained by a two time-scale update rule converge to a local nash
  equilibrium.
\newblock \emph{NeurIPS}, 30, 2017.

\bibitem[Ho and Salimans(2022)]{ho2022classifier}
Jonathan Ho and Tim Salimans.
\newblock Classifier-free diffusion guidance.
\newblock \emph{arXiv preprint arXiv:2207.12598}, 2022.

\bibitem[Ho et~al.(2020)Ho, Jain, and Abbeel]{ho2020denoising}
Jonathan Ho, Ajay Jain, and Pieter Abbeel.
\newblock Denoising diffusion probabilistic models.
\newblock \emph{NeurIPS}, 33:\penalty0 6840--6851, 2020.

\bibitem[Hu et~al.(2022)Hu, Shen, Wallis, Allen-Zhu, Li, Wang, Wang, and
  Chen]{hu2022lora}
Edward~J Hu, Yelong Shen, Phillip Wallis, Zeyuan Allen-Zhu, Yuanzhi Li, Shean
  Wang, Lu Wang, and Weizhu Chen.
\newblock Lo{RA}: Low-rank adaptation of large language models.
\newblock In \emph{ICLR}, 2022.

\bibitem[Huang et~al.(2016)Huang, Sun, Liu, Sedra, and
  Weinberger]{huang2016deep}
Gao Huang, Yu Sun, Zhuang Liu, Daniel Sedra, and Kilian~Q Weinberger.
\newblock Deep networks with stochastic depth.
\newblock In \emph{ECCV}, pages 646--661. Springer, 2016.

\bibitem[Huang et~al.(2020)Huang, Wang, Tai, Liu, Shen, Li, Li, and
  Huang]{huang2020curricularface}
Yuge Huang, Yuhan Wang, Ying Tai, Xiaoming Liu, Pengcheng Shen, Shaoxin Li,
  Jilin Li, and Feiyue Huang.
\newblock Curricularface: adaptive curriculum learning loss for deep face
  recognition.
\newblock In \emph{CVPR}, pages 5901--5910, 2020.

\bibitem[Jang et~al.(2016)Jang, Gu, and Poole]{jang2016categorical}
Eric Jang, Shixiang Gu, and Ben Poole.
\newblock Categorical reparameterization with gumbel-softmax.
\newblock \emph{arXiv preprint arXiv:1611.01144}, 2016.

\bibitem[Jang et~al.(2024)Jang, Jo, Lee, and Hwang]{jang2024identity}
Sangwon Jang, Jaehyeong Jo, Kimin Lee, and Sung~Ju Hwang.
\newblock Identity decoupling for multi-subject personalization of
  text-to-image models.
\newblock \emph{arXiv preprint arXiv:2404.04243}, 2024.

\bibitem[Kang et~al.(2023)Kang, Zhu, Zhang, Park, Shechtman, Paris, and
  Park]{kang2023scaling}
Minguk Kang, Jun-Yan Zhu, Richard Zhang, Jaesik Park, Eli Shechtman, Sylvain
  Paris, and Taesung Park.
\newblock Scaling up gans for text-to-image synthesis.
\newblock \emph{CVPR}, 2023.

\bibitem[Kim et~al.(2024)Kim, Lee, Joung, Kim, and Baek]{kim2024instantfamily}
Chanran Kim, Jeongin Lee, Shichang Joung, Bongmo Kim, and Yeul-Min Baek.
\newblock Instantfamily: Masked attention for zero-shot multi-id image
  generation.
\newblock \emph{arXiv preprint arXiv:2404.19427}, 2024.

\bibitem[Kong et~al.(2024)Kong, Zhang, Yang, Wang, Zhang, Wu, Chen, Liu, and
  Luo]{kong2024omg}
Zhe Kong, Yong Zhang, Tianyu Yang, Tao Wang, Kaihao Zhang, Bizhu Wu, Guanying
  Chen, Wei Liu, and Wenhan Luo.
\newblock Omg: Occlusion-friendly personalized multi-concept generation in
  diffusion models.
\newblock \emph{arXiv preprint arXiv:2403.10983}, 2024.

\bibitem[Kumari et~al.(2023{\natexlab{a}})Kumari, Zhang, Zhang, Shechtman, and
  Zhu]{kumari2022customdiffusion}
Nupur Kumari, Bingliang Zhang, Richard Zhang, Eli Shechtman, and Jun-Yan Zhu.
\newblock Multi-concept customization of text-to-image diffusion.
\newblock In \emph{CVPR}, 2023{\natexlab{a}}.

\bibitem[Kumari et~al.(2023{\natexlab{b}})Kumari, Zhang, Zhang, Shechtman, and
  Zhu]{kumari2022multi}
Nupur Kumari, Bingliang Zhang, Richard Zhang, Eli Shechtman, and Jun-Yan Zhu.
\newblock Multi-concept customization of text-to-image diffusion.
\newblock \emph{CVPR}, 2023{\natexlab{b}}.

\bibitem[Kumari et~al.(2023{\natexlab{c}})Kumari, Zhang, Zhang, Shechtman, and
  Zhu]{kumari2023multi}
Nupur Kumari, Bingliang Zhang, Richard Zhang, Eli Shechtman, and Jun-Yan Zhu.
\newblock Multi-concept customization of text-to-image diffusion.
\newblock In \emph{CVPR}, pages 1931--1941, 2023{\natexlab{c}}.

\bibitem[Kwon et~al.(2024)Kwon, Jenni, Li, Lee, Ye, and
  Heilbron]{kwon2024concept}
Gihyun Kwon, Simon Jenni, Dingzeyu Li, Joon-Young Lee, Jong~Chul Ye, and
  Fabian~Caba Heilbron.
\newblock Concept weaver: Enabling multi-concept fusion in text-to-image
  models.
\newblock In \emph{CVPR}, pages 8880--8889, 2024.

\bibitem[Li et~al.(2024{\natexlab{a}})Li, Li, and Hoi]{li2024blipdiffusion}
Dongxu Li, Junnan Li, and Steven Hoi.
\newblock Blip-diffusion: Pre-trained subject representation for controllable
  text-to-image generation and editing.
\newblock \emph{NeurIPS}, 36, 2024{\natexlab{a}}.

\bibitem[Li et~al.(2023)Li, Li, Savarese, and Hoi]{li2023blip}
Junnan Li, Dongxu Li, Silvio Savarese, and Steven Hoi.
\newblock Blip-2: Bootstrapping language-image pre-training with frozen image
  encoders and large language models.
\newblock \emph{arXiv preprint arXiv:2301.12597}, 2023.

\bibitem[Li et~al.(2017)Li, Bolkart, Black, Li, and Romero]{li2017learning}
Tianye Li, Timo Bolkart, Michael~J Black, Hao Li, and Javier Romero.
\newblock Learning a model of facial shape and expression from 4d scans.
\newblock \emph{ACM Trans. Graph.}, 36\penalty0 (6):\penalty0 194--1, 2017.

\bibitem[Li et~al.(2024{\natexlab{b}})Li, Cao, Wang, Qi, Cheng, and
  Shan]{li2024photomaker}
Zhen Li, Mingdeng Cao, Xintao Wang, Zhongang Qi, Ming-Ming Cheng, and Ying
  Shan.
\newblock Photomaker: Customizing realistic human photos via stacked id
  embedding.
\newblock In \emph{CVPR}, pages 8640--8650, 2024{\natexlab{b}}.

\bibitem[Liang et~al.(2024)Liang, Ma, Zhu, Deng, and Yang]{liang2024caphuman}
Chao Liang, Fan Ma, Linchao Zhu, Yingying Deng, and Yi Yang.
\newblock Caphuman: Capture your moments in parallel universes.
\newblock In \emph{CVPR}, pages 6400--6409, 2024.

\bibitem[Liu et~al.(2017)Liu, Wen, Yu, Li, Raj, and Song]{liu2017sphereface}
Weiyang Liu, Yandong Wen, Zhiding Yu, Ming Li, Bhiksha Raj, and Le Song.
\newblock Sphereface: Deep hypersphere embedding for face recognition.
\newblock In \emph{CVPR}, pages 212--220, 2017.

\bibitem[Liu et~al.(2015)Liu, Luo, Wang, and Tang]{liu2015deep}
Ziwei Liu, Ping Luo, Xiaogang Wang, and Xiaoou Tang.
\newblock Deep learning face attributes in the wild.
\newblock In \emph{ICCV}, pages 3730--3738, 2015.

\bibitem[Liu et~al.(2023)Liu, Zhang, Shen, Zheng, Zhu, Feng, Liu, Zhao, Zhou,
  and Cao]{liu2023cones}
Zhiheng Liu, Yifei Zhang, Yujun Shen, Kecheng Zheng, Kai Zhu, Ruili Feng, Yu
  Liu, Deli Zhao, Jingren Zhou, and Yang Cao.
\newblock Cones 2: Customizable image synthesis with multiple subjects.
\newblock \emph{arXiv preprint arXiv:2305.19327}, 2023.

\bibitem[Loshchilov and Hutter(2017)]{loshchilov2017decoupled}
Ilya Loshchilov and Frank Hutter.
\newblock Decoupled weight decay regularization.
\newblock \emph{arXiv preprint arXiv:1711.05101}, 2017.

\bibitem[Ma et~al.(2023)Ma, Liang, Chen, and Lu]{subject-diffusion}
Jian Ma, Junhao Liang, Chen Chen, and Haonan Lu.
\newblock Subject-diffusion: Open domain personalized text-to-image generation
  without test-time fine-tuning.
\newblock \emph{arXiv preprint arXiv:2307.11410}, 2023.

\bibitem[Ostashev et~al.(2024)Ostashev, Fang, Tulyakov, Aberman,
  et~al.]{ostashev2024moa}
Daniil Ostashev, Yuwei Fang, Sergey Tulyakov, Kfir Aberman, et~al.
\newblock Moa: Mixture-of-attention for subject-context disentanglement in
  personalized image generation.
\newblock \emph{arXiv preprint arXiv:2404.11565}, 2024.

\bibitem[Peng et~al.(2024)Peng, Zhu, Jiang, Tai, Luo, Zhang, Lin, Jin, Wang,
  and Ji]{peng2024portraitbooth}
Xu Peng, Junwei Zhu, Boyuan Jiang, Ying Tai, Donghao Luo, Jiangning Zhang, Wei
  Lin, Taisong Jin, Chengjie Wang, and Rongrong Ji.
\newblock Portraitbooth: A versatile portrait model for fast identity-preserved
  personalization.
\newblock In \emph{CVPR}, pages 27080--27090, 2024.

\bibitem[Podell et~al.(2023)Podell, English, Lacey, Blattmann, Dockhorn,
  M{\"u}ller, Penna, and Rombach]{podell2023sdxl}
Dustin Podell, Zion English, Kyle Lacey, Andreas Blattmann, Tim Dockhorn, Jonas
  M{\"u}ller, Joe Penna, and Robin Rombach.
\newblock Sdxl: Improving latent diffusion models for high-resolution image
  synthesis.
\newblock \emph{arXiv preprint arXiv:2307.01952}, 2023.

\bibitem[Radford et~al.(2015)Radford, Metz, and
  Chintala]{radford2015unsupervised}
Alec Radford, Luke Metz, and Soumith Chintala.
\newblock Unsupervised representation learning with deep convolutional
  generative adversarial networks.
\newblock \emph{arXiv preprint arXiv:1511.06434}, 2015.

\bibitem[Radford et~al.(2021)Radford, Kim, Hallacy, Ramesh, Goh, Agarwal,
  Sastry, Askell, Mishkin, Clark, et~al.]{radford2021learning}
Alec Radford, Jong~Wook Kim, Chris Hallacy, Aditya Ramesh, Gabriel Goh,
  Sandhini Agarwal, Girish Sastry, Amanda Askell, Pamela Mishkin, Jack Clark,
  et~al.
\newblock Learning transferable visual models from natural language
  supervision.
\newblock In \emph{ICML}, pages 8748--8763. PMLR, 2021.

\bibitem[Ramesh et~al.(2022)Ramesh, Dhariwal, Nichol, Chu, and
  Chen]{ramesh2022hierarchical}
Aditya Ramesh, Prafulla Dhariwal, Alex Nichol, Casey Chu, and Mark Chen.
\newblock Hierarchical text-conditional image generation with clip latents.
\newblock \emph{arXiv preprint arXiv:2204.06125}, 2022.

\bibitem[Rombach et~al.(2022)Rombach, Blattmann, Lorenz, Esser, and Ommer]{ldm}
Robin Rombach, Andreas Blattmann, Dominik Lorenz, Patrick Esser, and Bj{\"o}rn
  Ommer.
\newblock High-resolution image synthesis with latent diffusion models.
\newblock In \emph{CVPR}, pages 10684--10695, 2022.

\bibitem[Ruiz et~al.(2023{\natexlab{a}})Ruiz, Li, Jampani, Pritch, Rubinstein,
  and Aberman]{ruiz2023dreambooth}
Nataniel Ruiz, Yuanzhen Li, Varun Jampani, Yael Pritch, Michael Rubinstein, and
  Kfir Aberman.
\newblock Dreambooth: Fine tuning text-to-image diffusion models for
  subject-driven generation.
\newblock In \emph{CVPR}, pages 22500--22510, 2023{\natexlab{a}}.

\bibitem[Ruiz et~al.(2023{\natexlab{b}})Ruiz, Li, Jampani, Wei, Hou, Pritch,
  Wadhwa, Rubinstein, and Aberman]{ruiz2023hyperdreambooth}
Nataniel Ruiz, Yuanzhen Li, Varun Jampani, Wei Wei, Tingbo Hou, Yael Pritch,
  Neal Wadhwa, Michael Rubinstein, and Kfir Aberman.
\newblock Hyperdreambooth: Hypernetworks for fast personalization of
  text-to-image models.
\newblock \emph{arXiv preprint arXiv:2307.06949}, 2023{\natexlab{b}}.

\bibitem[Saharia et~al.(2022)Saharia, Chan, Saxena, Li, Whang, Denton,
  Ghasemipour, Gontijo~Lopes, Karagol~Ayan, Salimans,
  et~al.]{saharia2022photorealistic}
Chitwan Saharia, William Chan, Saurabh Saxena, Lala Li, Jay Whang, Emily~L
  Denton, Kamyar Ghasemipour, Raphael Gontijo~Lopes, Burcu Karagol~Ayan, Tim
  Salimans, et~al.
\newblock Photorealistic text-to-image diffusion models with deep language
  understanding.
\newblock In \emph{NeurIPS}, pages 36479--36494, 2022.

\bibitem[Sauer et~al.(2023)Sauer, Karras, Laine, Geiger, and
  Aila]{sauer2023stylegan}
Axel Sauer, Tero Karras, Samuli Laine, Andreas Geiger, and Timo Aila.
\newblock Stylegan-t: Unlocking the power of gans for fast large-scale
  text-to-image synthesis.
\newblock In \emph{ICML}, pages 30105--30118. PMLR, 2023.

\bibitem[Schroff et~al.(2015)Schroff, Kalenichenko, and
  Philbin]{schroff2015facenet}
Florian Schroff, Dmitry Kalenichenko, and James Philbin.
\newblock Facenet: A unified embedding for face recognition and clustering.
\newblock In \emph{CVPR}, pages 815--823, 2015.

\bibitem[Schuhmann(2022)]{LAION-Aesthetics}
Christoph Schuhmann.
\newblock Laion-aesthetics.
\newblock \url{https://laion.ai/blog/laion-aesthetics/}, 2022.

\bibitem[Schuhmann et~al.(2021)Schuhmann, Vencu, Beaumont, Kaczmarczyk, Mullis,
  Katta, Coombes, Jitsev, and Komatsuzaki]{schuhmann2021laion}
Christoph Schuhmann, Richard Vencu, Romain Beaumont, Robert Kaczmarczyk,
  Clayton Mullis, Aarush Katta, Theo Coombes, Jenia Jitsev, and Aran
  Komatsuzaki.
\newblock Laion-400m: Open dataset of clip-filtered 400 million image-text
  pairs.
\newblock \emph{arXiv preprint arXiv:2111.02114}, 2021.

\bibitem[Schuhmann et~al.(2022)Schuhmann, Beaumont, Vencu, Gordon, Wightman,
  Cherti, Coombes, Katta, Mullis, Wortsman, et~al.]{schuhmann2022laion}
Christoph Schuhmann, Romain Beaumont, Richard Vencu, Cade Gordon, Ross
  Wightman, Mehdi Cherti, Theo Coombes, Aarush Katta, Clayton Mullis, Mitchell
  Wortsman, et~al.
\newblock Laion-5b: An open large-scale dataset for training next generation
  image-text models.
\newblock In \emph{NeurIPS}, pages 25278--25294, 2022.

\bibitem[Sohl-Dickstein et~al.(2015)Sohl-Dickstein, Weiss, Maheswaranathan, and
  Ganguli]{sohl2015deep}
Jascha Sohl-Dickstein, Eric Weiss, Niru Maheswaranathan, and Surya Ganguli.
\newblock Deep unsupervised learning using nonequilibrium thermodynamics.
\newblock In \emph{ICML}, pages 2256--2265, 2015.

\bibitem[Song et~al.(2020)Song, Meng, and Ermon]{song2020denoising}
Jiaming Song, Chenlin Meng, and Stefano Ermon.
\newblock Denoising diffusion implicit models.
\newblock \emph{arXiv preprint arXiv:2010.02502}, 2020.

\bibitem[Song and Ermon(2019)]{song2019generative}
Yang Song and Stefano Ermon.
\newblock Generative modeling by estimating gradients of the data distribution.
\newblock \emph{NeurIPS}, 32, 2019.

\bibitem[Wang et~al.(2018)Wang, Wang, Zhou, Ji, Gong, Zhou, Li, and
  Liu]{wang2018cosface}
Hao Wang, Yitong Wang, Zheng Zhou, Xing Ji, Dihong Gong, Jingchao Zhou, Zhifeng
  Li, and Wei Liu.
\newblock Cosface: Large margin cosine loss for deep face recognition.
\newblock In \emph{CVPR}, pages 5265--5274, 2018.

\bibitem[Wang et~al.(2024)Wang, Bai, Wang, Qin, and Chen]{wang2024instantid}
Qixun Wang, Xu Bai, Haofan Wang, Zekui Qin, and Anthony Chen.
\newblock Instantid: Zero-shot identity-preserving generation in seconds.
\newblock \emph{arXiv preprint arXiv:2401.07519}, 2024.

\bibitem[Wang et~al.(2023)Wang, Wang, Xie, Qi, Shan, Wang, and
  Luo]{wang2023styleadapter}
Zhouxia Wang, Xintao Wang, Liangbin Xie, Zhongang Qi, Ying Shan, Wenping Wang,
  and Ping Luo.
\newblock Styleadapter: A single-pass lora-free model for stylized image
  generation.
\newblock \emph{arXiv preprint arXiv:2309.01770}, 2023.

\bibitem[Wei et~al.(2023)Wei, Zhang, Ji, Bai, Zhang, and Zuo]{wei2023elite}
Yuxiang Wei, Yabo Zhang, Zhilong Ji, Jinfeng Bai, Lei Zhang, and Wangmeng Zuo.
\newblock Elite: Encoding visual concepts into textual embeddings for
  customized text-to-image generation.
\newblock \emph{arXiv preprint arXiv:2302.13848}, 2023.

\bibitem[Wu et~al.(2024)Wu, Li, Zheng, Wang, and Li]{wu2024infinite}
Yi Wu, Ziqiang Li, Heliang Zheng, Chaoyue Wang, and Bin Li.
\newblock Infinite-id: Identity-preserved personalization via id-semantics
  decoupling paradigm.
\newblock \emph{arXiv preprint arXiv:2403.11781}, 2024.

\bibitem[Xiao et~al.(2023{\natexlab{a}})Xiao, Yin, Freeman, Durand, and
  Han]{fastcomposer}
Guangxuan Xiao, Tianwei Yin, William~T Freeman, Fr{\'e}do Durand, and Song Han.
\newblock Fastcomposer: Tuning-free multi-subject image generation with
  localized attention.
\newblock \emph{arXiv preprint arXiv:2305.10431}, 2023{\natexlab{a}}.

\bibitem[Xiao et~al.(2023{\natexlab{b}})Xiao, Yin, Freeman, Durand, and
  Han]{xiao2023fastcomposer}
Guangxuan Xiao, Tianwei Yin, William~T Freeman, Fr{\'e}do Durand, and Song Han.
\newblock Fastcomposer: Tuning-free multi-subject image generation with
  localized attention.
\newblock \emph{arXiv preprint arXiv:2305.10431}, 2023{\natexlab{b}}.

\bibitem[Yan et~al.(2023)Yan, Zhang, Wang, Zhou, Zhang, Cheng, Yu, and
  Fu]{yan2023facestudio}
Yuxuan Yan, Chi Zhang, Rui Wang, Yichao Zhou, Gege Zhang, Pei Cheng, Gang Yu,
  and Bin Fu.
\newblock Facestudio: Put your face everywhere in seconds.
\newblock \emph{arXiv preprint arXiv:2312.02663}, 2023.

\bibitem[Ye et~al.(2023)Ye, Zhang, Liu, Han, and Yang]{ye2023ip-adapter}
Hu Ye, Jun Zhang, Sibo Liu, Xiao Han, and Wei Yang.
\newblock Ip-adapter: Text compatible image prompt adapter for text-to-image
  diffusion models.
\newblock \emph{arXiv preprint arXiv:2308.06721}, 2023.

\bibitem[Zhang(2023)]{refonly}
Lvmin Zhang.
\newblock Reference-only controlnet.
\newblock
  \url{https://github.com/Mikubill/sd-webui-controlnet/discussions/1236}, 2023.

\bibitem[Zhang et~al.(2023)Zhang, Rao, and Agrawala]{zhang2023adding}
Lvmin Zhang, Anyi Rao, and Maneesh Agrawala.
\newblock Adding conditional control to text-to-image diffusion models.
\newblock In \emph{ICCV}, pages 3836--3847, 2023.

\bibitem[Zhang et~al.(2024)Zhang, Huang, Chen, Zhang, Wu, Feng, Wang, Shen,
  Liu, and Luo]{zhang2024flashface}
Shilong Zhang, Lianghua Huang, Xi Chen, Yifei Zhang, Zhi-Fan Wu, Yutong Feng,
  Wei Wang, Yujun Shen, Yu Liu, and Ping Luo.
\newblock Flashface: Human image personalization with high-fidelity identity
  preservation.
\newblock \emph{arXiv preprint arXiv:2403.17008}, 2024.

\end{thebibliography}
}

\clearpage
\appendix
\begin{center}{\bf \Large Appendix}\end{center}\vspace{-2mm}
\renewcommand{\thetable}{\Roman{table}}
\renewcommand{\thefigure}{\Roman{figure}}
\setcounter{table}{0}
\setcounter{figure}{0}

\section{Face Structure Scale}
\label{app:A}

\begin{figure}[b]
	\centering
	\includegraphics[width=1.0\linewidth]{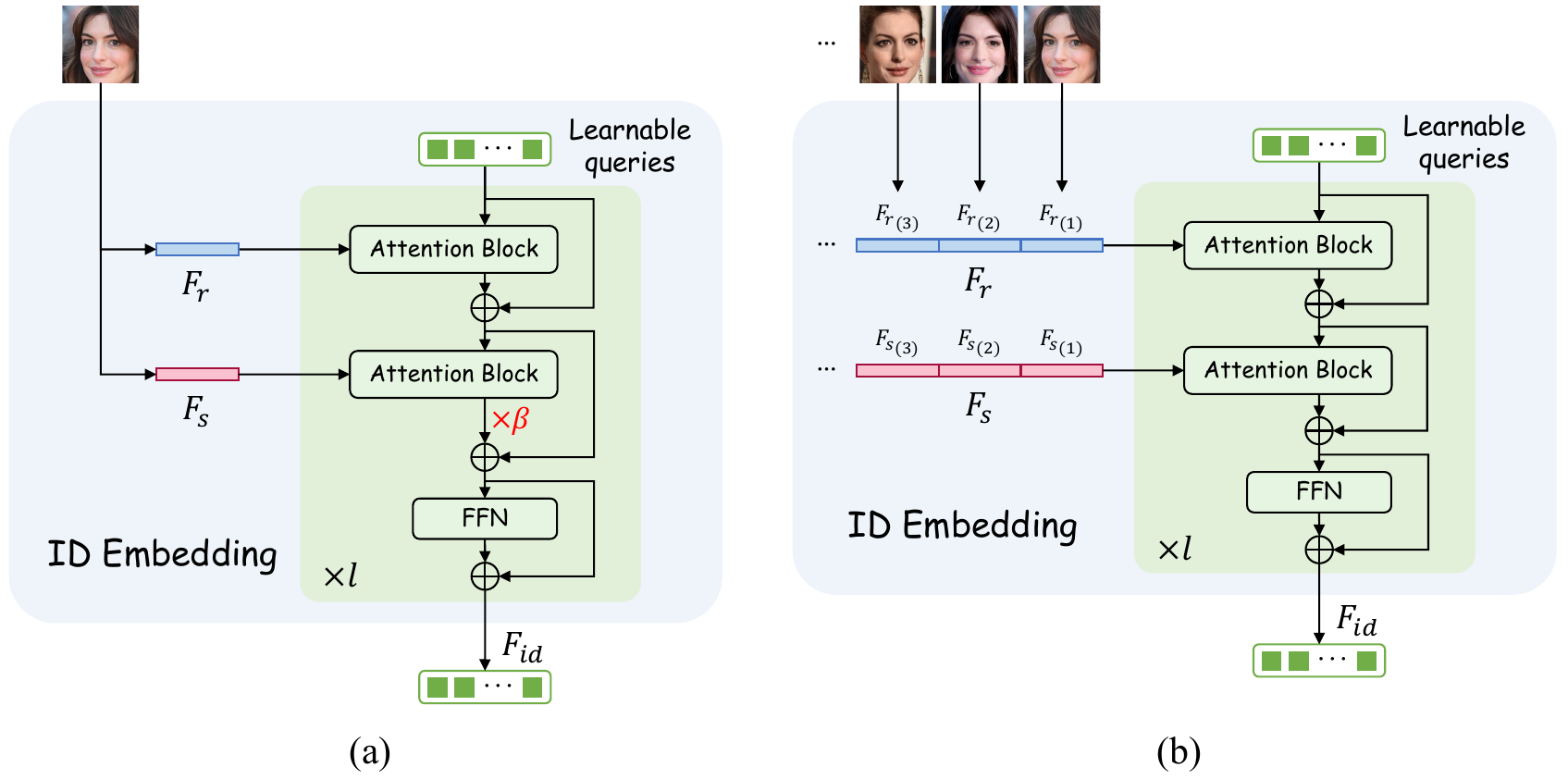}
	\caption{\textbf{Illustrations of (a) face structure scale $\beta$ and (b) single-ID multi-reference embedding in the inference phrase.}}
	\vspace{-3mm}
	\label{fig:face_structure_multi_refs}
\end{figure}

\begin{figure*}[bp]
	\centering
	\includegraphics[width=1.0\linewidth]{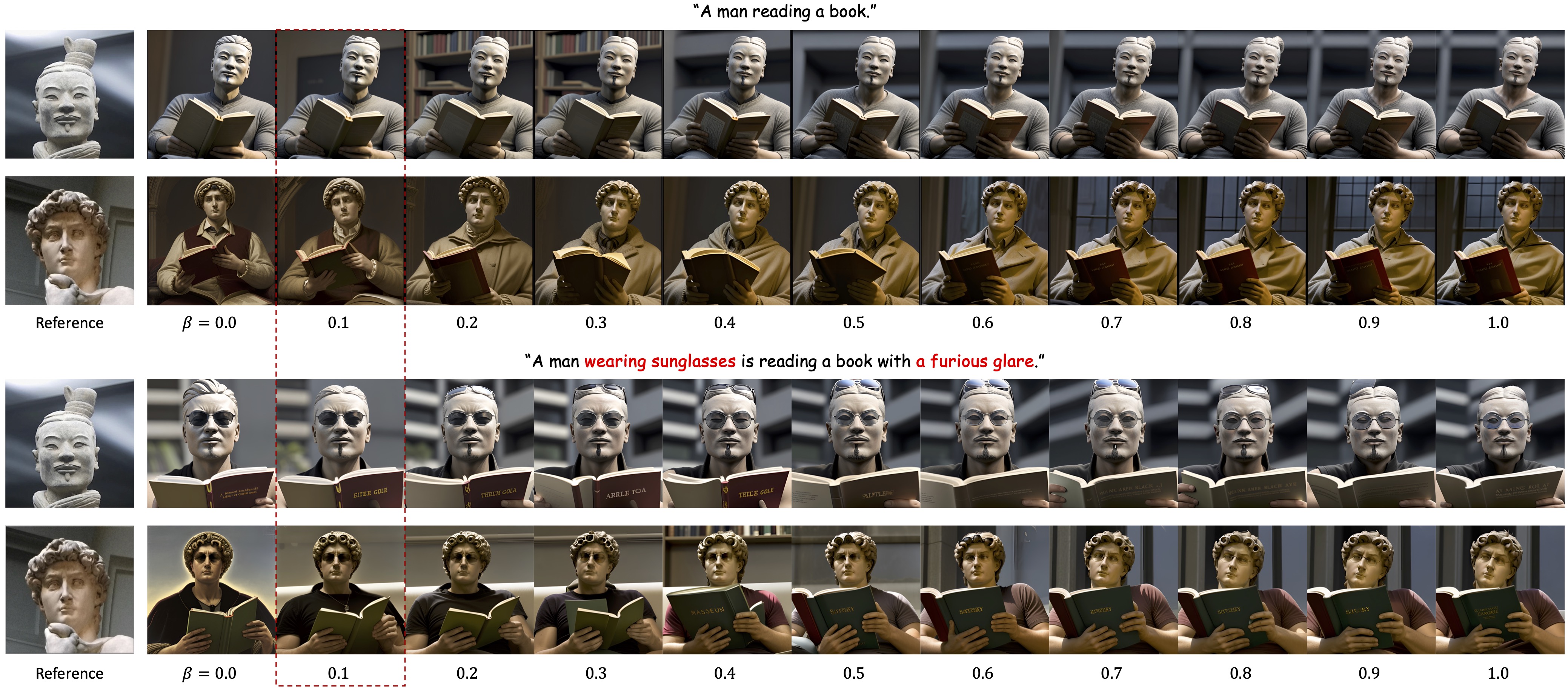}
	\caption{\textbf{The effect of the face structure scale.}}
	\vspace{-5mm}
	\label{fig:face_structure}
\end{figure*}

As illustrated in \tabref{tab:single-id-ablation} of the main paper, the introduction of the face structure branch has indeed enhanced ID similarity. However, it has also compromised some aspects of prompt consistency. In practical implementation, a scaling factor $\beta$ (see \figref{fig:face_structure_multi_refs}(a)) can be incorporated to modulate the intensity of the face structure branch, thereby enabling a flexible trade-off between face similarity and prompt consistency. \figref{fig:face_structure} provides a visualization of the impact of different $\beta$ values on the generated outcomes. As expected, a larger $\beta$ value leads to increased facial shape and texture fidelity (attributed to the detailed spatial information in the face structure branch), but also results in decreased facial editability (\eg, expression editing, due to the insufficient disentanglement of face structure features). Empirically, a lower $\beta$ value (\eg, 0.1) can be selected to achieve a balance between prompt consistency and ID alignment.

\section{Single-ID Multi-Reference Generation}
\label{app:B}

As described in \secref{sec:3.2} of the main paper, our method can seamlessly extend to the generation of multi-reference images for a single ID (see \figref{fig:face_structure_multi_refs}(b)). \tabref{tab:num_references} provides a quantitative assessment of the advantages of using multi-reference images.  Here, we continue to use the 15 IDs described in \secref{sec:4.2} for evaluation, but instead of using the same reference image to compute face similarity, we use a different image of the same ID as the target. The findings demonstrate that multi-reference images can improve ID similarity without compromising prompt consistency, FID, and aesthetic scores.

\section{Visualization of Multi-ID Routing Map}
\label{app:C}

\begin{figure*}[t]
	\centering
	\includegraphics[width=1.0\linewidth]{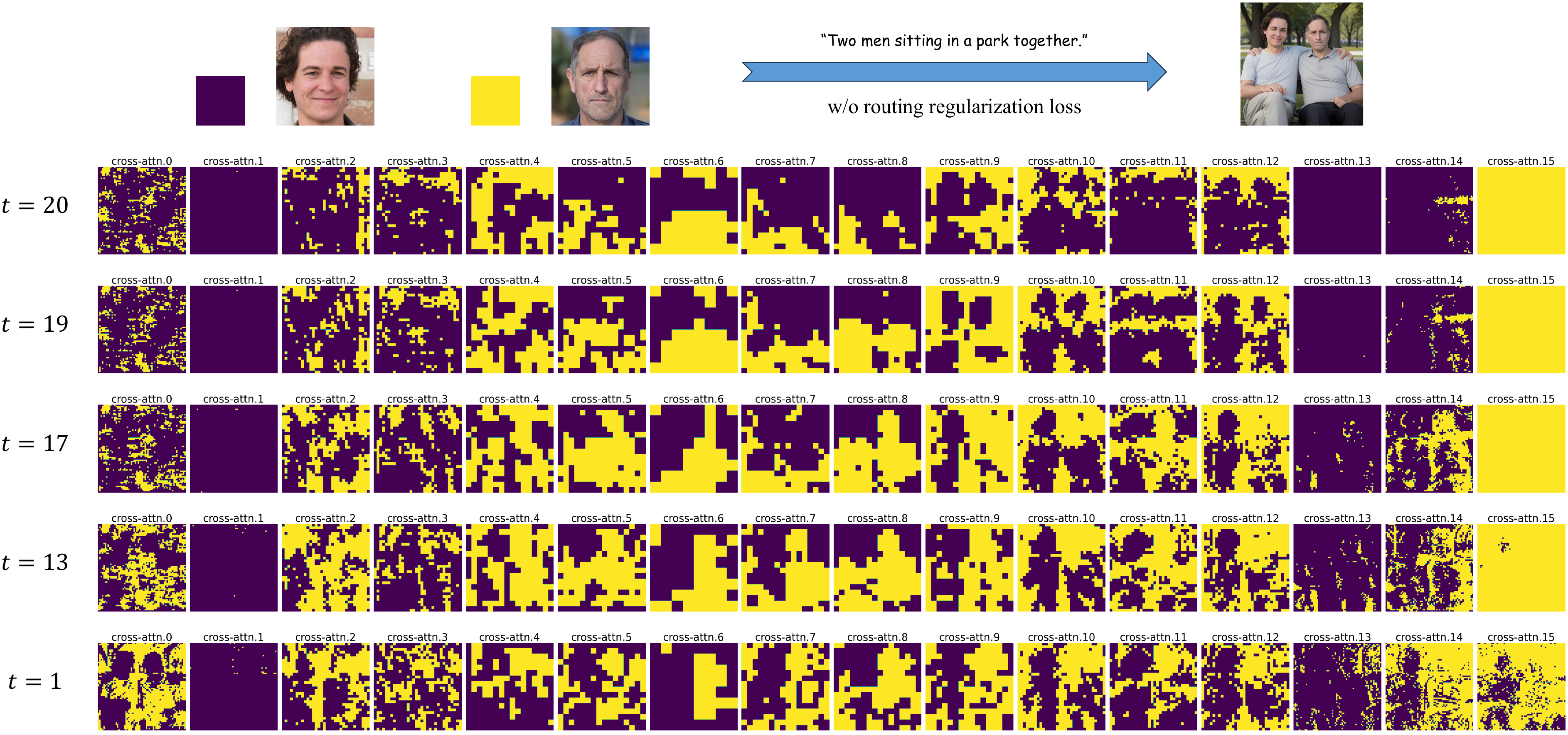}
	\caption{\textbf{Visualization of the routing map within each cross-attention layer of the U-Net without the routing regularization loss.}}
	\label{fig:routing_map_1}
\end{figure*}

\begin{figure*}[t]
	\centering
	\includegraphics[width=1.0\linewidth]{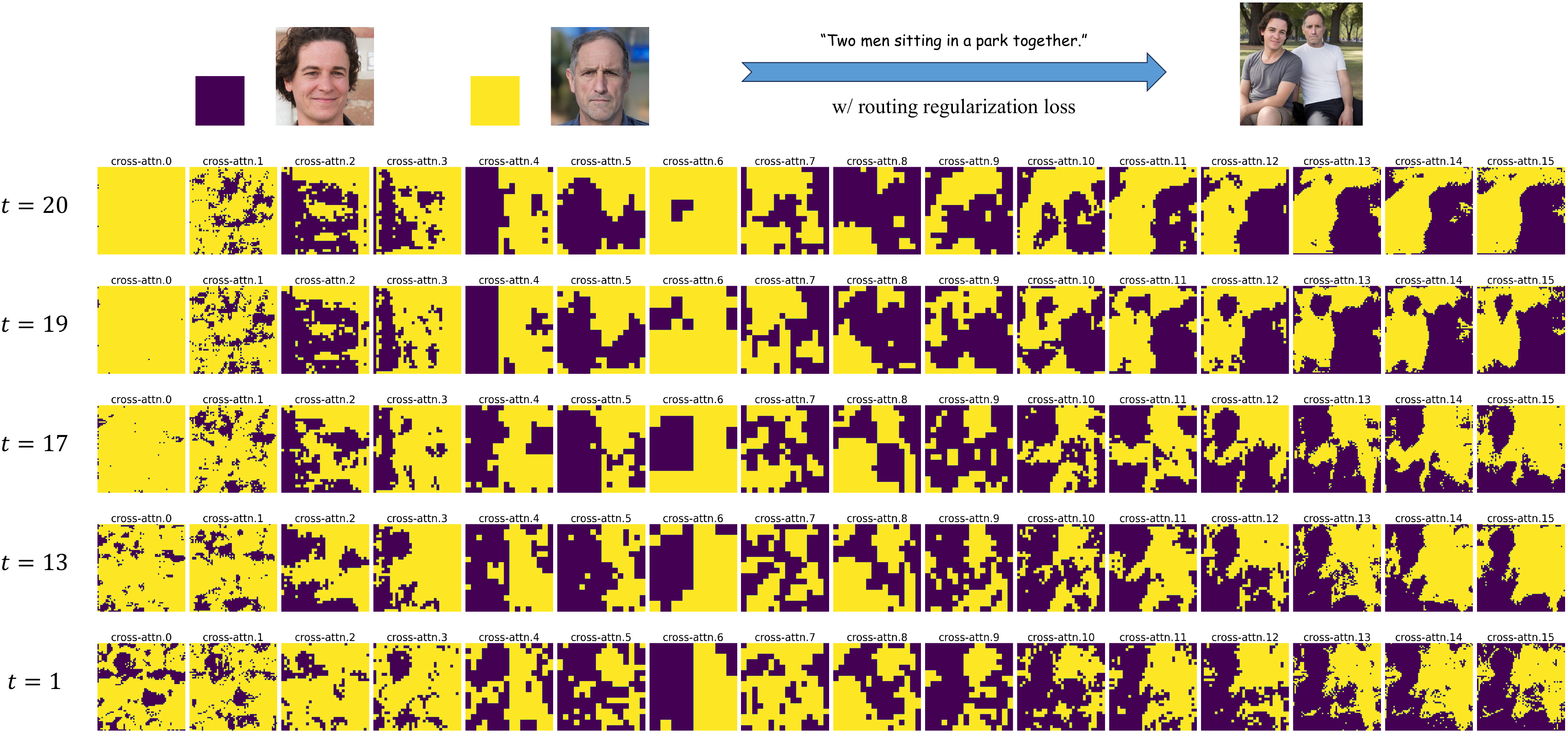}
	\caption{\textbf{Visualization of the routing map within each cross-attention layer of the U-Net with the routing regularization loss.}}
	\label{fig:routing_map_2}
\end{figure*}

\figref{fig:routing_map_1} and \figref{fig:routing_map_2} display the routing map of each cross-attention layer of the U-Net under different diffusion steps. It is observed that, with the utilization of routing regularization loss, the ID router can discern different IDs at earlier time steps and in a more pronounced manner, particularly within the cross-attention layers of the U-Net's decoder.

\begin{table}[t]
	\centering
	\tablestyle{4pt}{1.2}\scriptsize\begin{tabular}{x{86} | x{18} x{32}  x{16}  x{24}}
		\multirow{2}*{Number of reference images} &  \multirow{2}*{\shortstack{Face Sim. \\$\uparrow$ (\%) }} & \multirow{2}*{\shortstack{CLIP-T \\ $\uparrow$ (\%)}} & \multirow{2}*{FID $\downarrow$} & \multirow{2}*{\shortstack{LAION-\\Aes $\uparrow$}} \\
		&&&&\\
		\shline
		1  &  55.5 & \textbf{26.1} & 123.4 & 6.42 \\
		2 & 58.5 & 26.1 & \textbf{122.8} & 6.41 \\	
		4 & \textbf{61.2} & 25.9 & 124.7 & \textbf{6.44} \\	
	\end{tabular}
	\caption{\textbf{Performance of using multi-reference images.} }
	\label{tab:num_references}
\end{table}

\section{More Example Generations}
\label{app:D}

\begin{figure*}[t]
	\centering
	\includegraphics[width=1.0\linewidth]{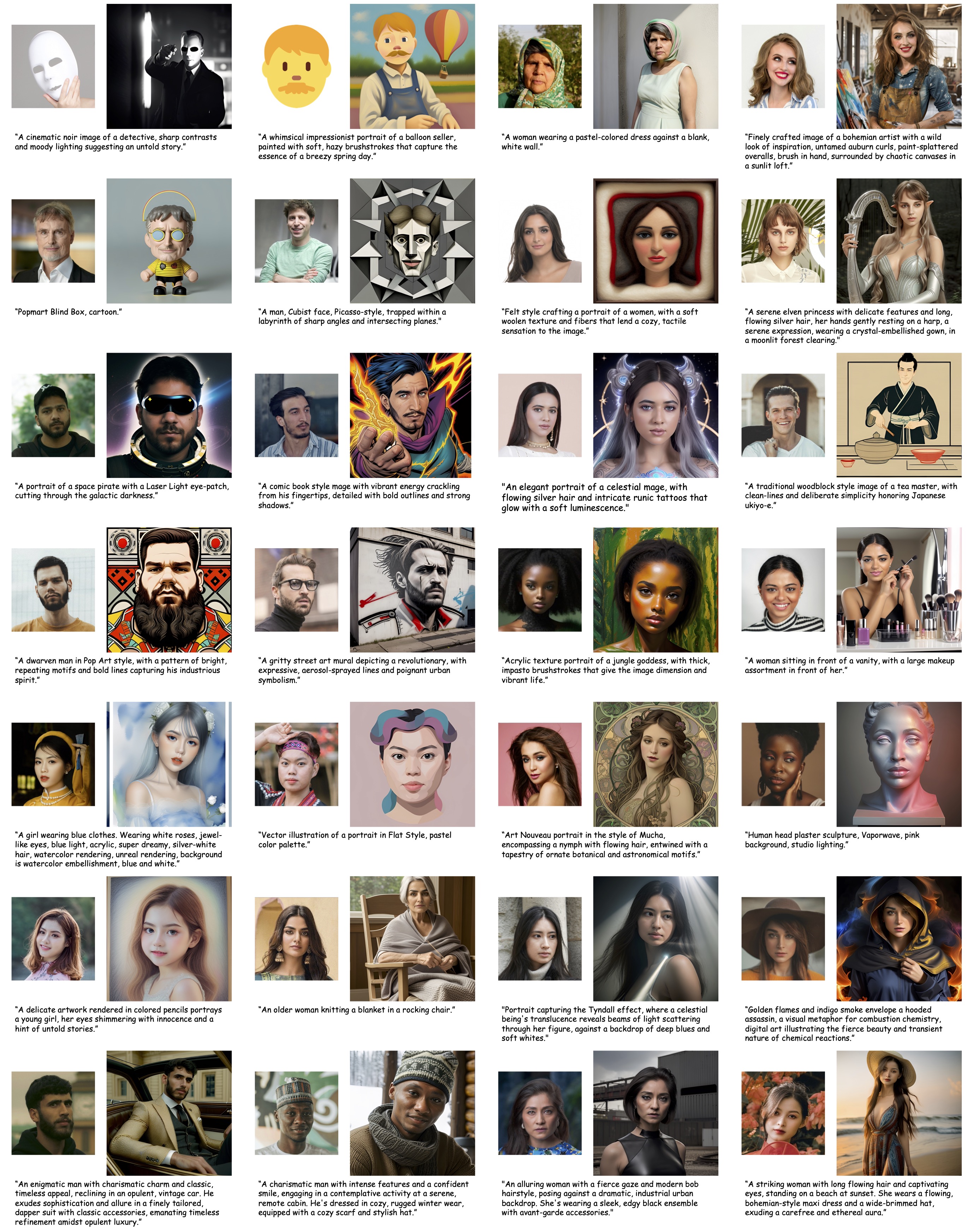}
	\caption{\textbf{Text-to-single-ID personalization examples.}}
	\label{fig:examples_1}
\end{figure*}

\begin{figure*}[t]
	\centering
	\includegraphics[width=1.0\linewidth]{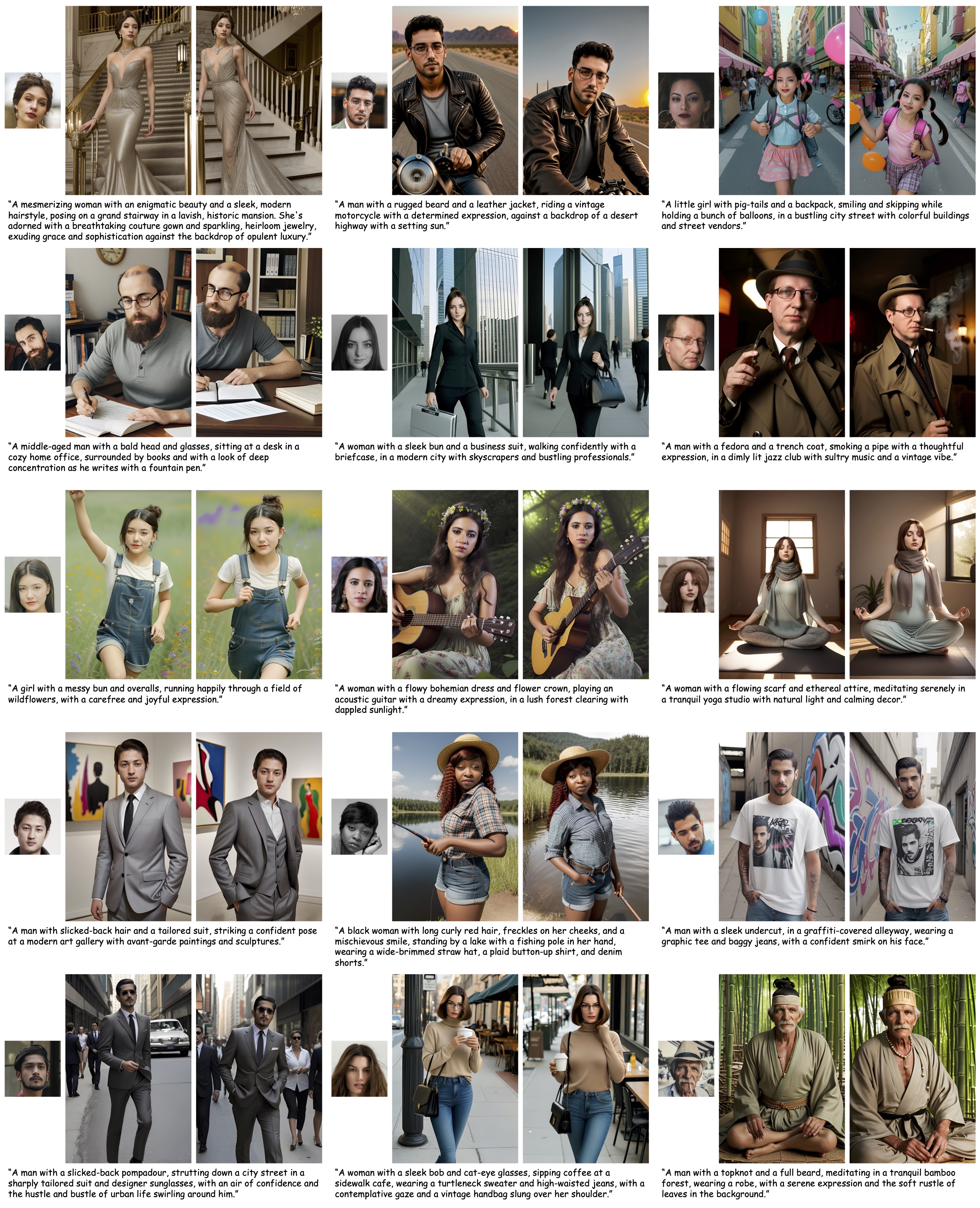}
	\caption{\textbf{Text-to-single-ID personalization examples.}}
	\label{fig:examples_2}
\end{figure*}

\begin{figure*}[t]
	\centering
	\includegraphics[width=1.0\linewidth]{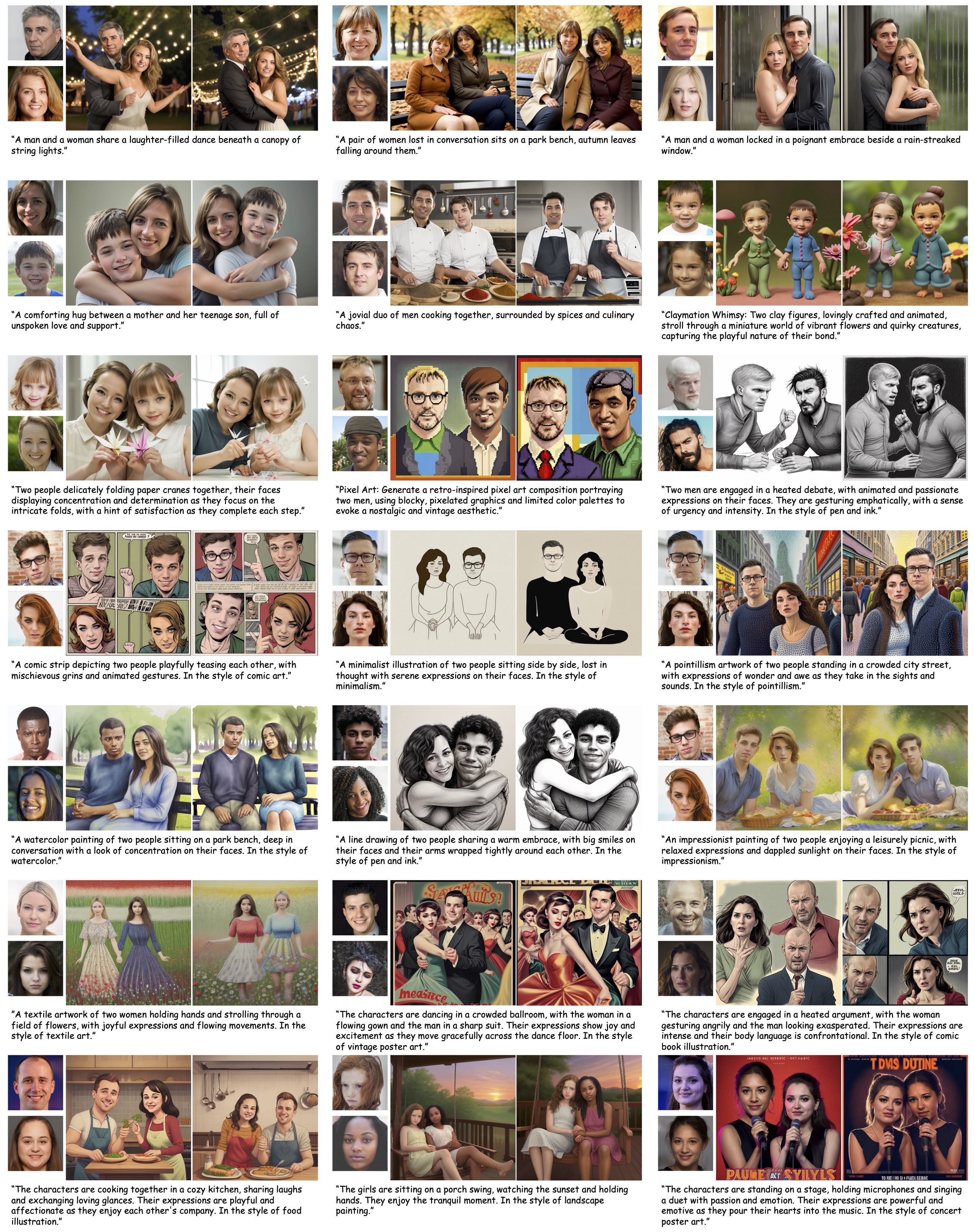}
	\caption{\textbf{Text-to-multi-ID personalization examples.}}
	\label{fig:examples_3}
\end{figure*}

\begin{figure*}[t]
	\centering
	\includegraphics[width=1.0\linewidth]{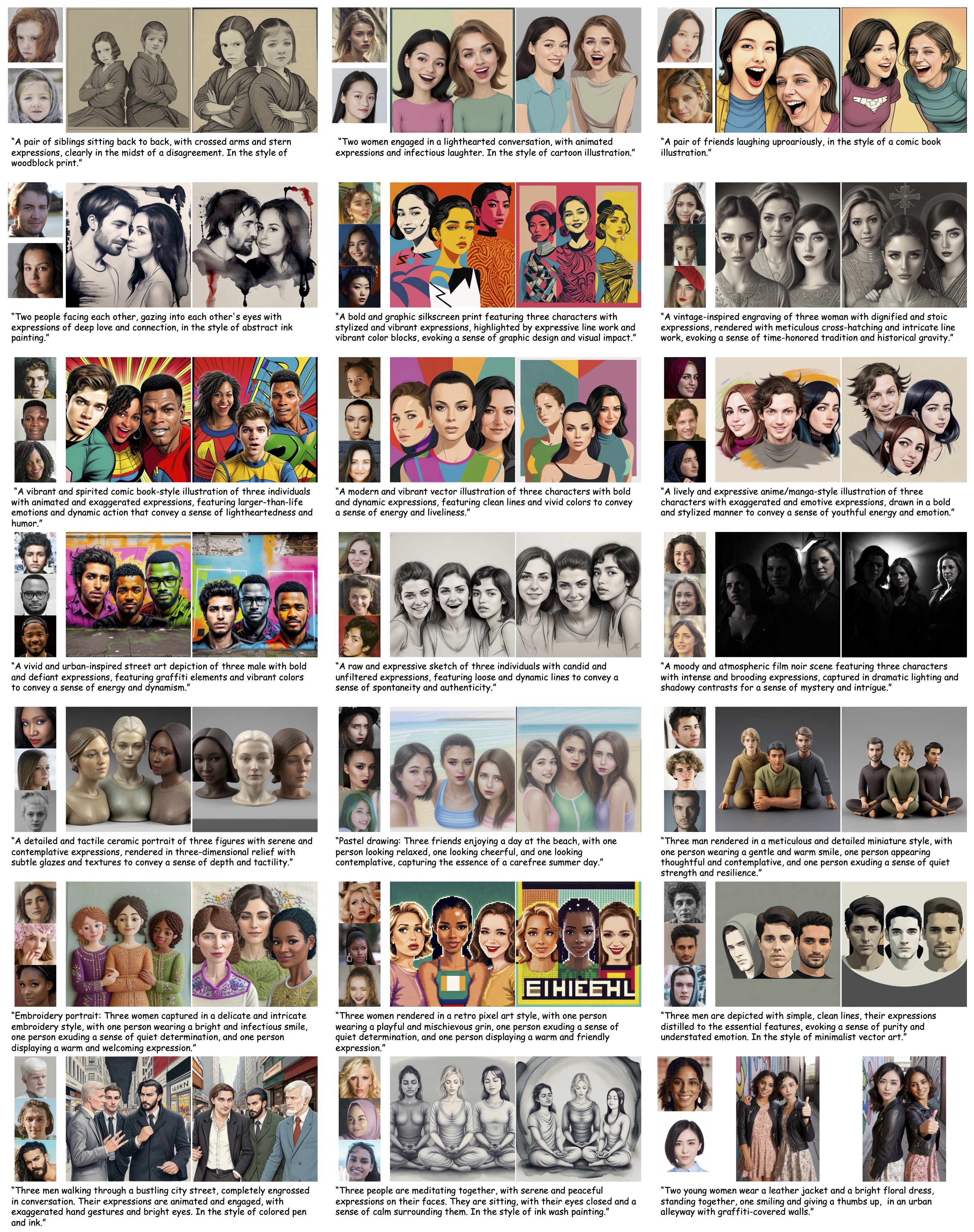}
	\caption{\textbf{Text-to-multi-ID personalization examples.}}
	\label{fig:examples_4}
\end{figure*}

\figref{fig:examples_1}-\ref{fig:examples_4} present more personalized human image results of UniPortrait, demonstrating the outstanding performance of our approach in single- and multi-ID customization again.

\section{Open-Sourced Text-to-Image Models}
\label{app:E}

\tabref{tab:urls} shows the URLs of the utilized open-sourced text-to-image models and LoRAs in this paper.

\begin{table}[t]
	\centering
	\tablestyle{1pt}{1.2}\scriptsize\begin{tabular}{y{50} | y{180}}
		Model / LoRA &  URL\\
		\shline
		Realistic Vision V4.0  &  https://huggingface.co/SG161222/Realistic\_Vision\_V4.0\_noVAE \\
		\hline
		ToonYou & https://civitai.com/models/30240/toonyou \\
		\hline
		disney-pixar-cartoon-typeb & https://civitai.com/models/75650/disney-pixar-cartoon-typeb \\
		\hline
		chosen-chinese-stylensfw-hentai & https://civitai.com/models/95643/chosen-chinese-stylensfw-hentai
	\end{tabular}
	\caption{\textbf{URLs of the used models and LoRAs in this paper.} }
	\label{tab:urls}
\end{table}

\end{document}